\pdfoutput=1
\documentclass[dvipsnames]{article} %
\usepackage[preprint]{neurips_2025}

\usepackage{amsmath,amsfonts,bm}

\def\eqref#1{equation~\ref{#1}}

\def\1{\bm{1}}

\DeclareMathAlphabet{\mathsfit}{\encodingdefault}{\sfdefault}{m}{sl}
\SetMathAlphabet{\mathsfit}{bold}{\encodingdefault}{\sfdefault}{bx}{n}

\usepackage{hyperref}
\usepackage{url}
\usepackage{booktabs}%
\usepackage{changepage} %
\usepackage{acronym}

\usepackage{xspace}
\usepackage{graphicx} %

\usepackage{cleveref}
\usepackage{enumitem}
\usepackage{tabularx}
\usepackage{multirow}

\usepackage{pifont}     %
\usepackage{array}      %

\newcommand{\cgreenmark}{\textcolor{green!60!black}{\ding{51}}} %
\newcommand{\xredmark}{\textcolor{red}{\ding{55}}}            %

\newcommand{\ours}
{\textsc{X-Reasoner}\xspace}
\newcommand{\oursmed}
{\textsc{X-Reasoner-Med}\xspace}
\renewcommand{\vec}{\mathbf}
\usepackage[most,skins,theorems]{tcolorbox}
\usepackage{tikz}

\tcbset{
  aibox/.style={
    width=\linewidth,
    top=8pt,
    bottom=4pt,
    colback=blue!6!white,
    colframe=black,
    colbacktitle=black,
    enhanced,
    center,
    attach boxed title to top left={yshift=-0.1in,xshift=0.15in},
    boxed title style={boxrule=0pt,colframe=white,},
  }
}

\newtcolorbox{AIbox}[2][]{aibox,title=#2,#1}

\newlength\savewidth\newcommand\shline{\noalign{\global\savewidth\arrayrulewidth
  \global\arrayrulewidth 1pt}\hline\noalign{\global\arrayrulewidth\savewidth}}
\newcommand{\tablestyle}[2]{\setlength{\tabcolsep}{#1}\renewcommand{\arraystretch}{#2}\centering\footnotesize}

\usepackage{amssymb}
\definecolor{myblue}{HTML}{191970}
\definecolor{myred}{HTML}{B22222}

\newcommand{\bluesquare}{%
  \textcolor{myblue}{$\blacksquare$}%
}
\newcommand{\redsquare}{%
  \textcolor{myred}{$\blacksquare$}%
}

\newcommand{\icon}[2][1em]{%
  \raisebox{-.15\height}{\includegraphics[height=#1]{#2}}%
}

\newcommand{\base}{\texttt{Qwen2.5-VL-7B-Instruct}}

\usepackage{subcaption}

\usepackage[acronym]{glossaries}
\glsdisablehyper

\newacronym{SFT}{SFT}{Supervised Fine-Tuning}
\newacronym{RL}{RL}{Reinforcement Learning}

\title{\ours: Towards Generalizable Reasoning Across Modalities and Domains}

\author{
\textbf{Qianchu Liu\thanks{~~Equal contributions.}~\,, Sheng Zhang$^{*}$, Guanghui Qin$^{*}$, Timothy Ossowski,
Yu Gu, Ying Jin}\\
\textbf{Sid Kiblawi, Sam Preston, Mu Wei, Paul Vozila, Tristan Naumann, Hoifung Poon}\vspace{0.3em}\\
Microsoft Research\vspace{0.3em}\\
\href{https://github.com/microsoft/x-reasoner}{\textcolor{Blue}{\bf \texttt{github.com/microsoft/x-reasoner}}}
}
\date{}

\begin{document}

\maketitle

\begin{abstract}
Recent proprietary models (e.g., o3) have begun to demonstrate strong multimodal reasoning capabilities. Yet, most existing open-source research concentrates on training text-only reasoning models, with evaluations limited to mainly mathematical and general-domain tasks. Therefore, it remains unclear how to effectively extend reasoning capabilities beyond text input and general domains. This paper explores a fundamental research question: \emph{Is reasoning generalizable across modalities and domains?} Our findings support an affirmative answer: 
\emph{General-domain text-based post-training can enable such strong generalizable reasoning.} Leveraging this finding, we introduce \ours, a vision-language model post-trained solely on general-domain text for generalizable reasoning, using a two-stage approach: an initial supervised fine-tuning phase with distilled long chain-of-thoughts, followed by reinforcement learning with verifiable rewards. Experiments show that \ours successfully transfers reasoning capabilities to both multimodal and out-of-domain settings, outperforming existing state-of-the-art models trained with in-domain and multimodal data across various general and medical benchmarks (\Cref{fig:overview_figure}).
Additionally, we find that \textsc{X-Reasoner}’s performance in specialized domains can be further enhanced through continued training on domain-specific text-only data. Building upon this, we introduce \oursmed, a medical-specialized variant that achieves new state of the art on numerous text-only and multimodal medical benchmarks.

\begin{figure}[!ht]
    \centering
        \vspace{-2pt}
    \includegraphics[width=1.0\textwidth]
    {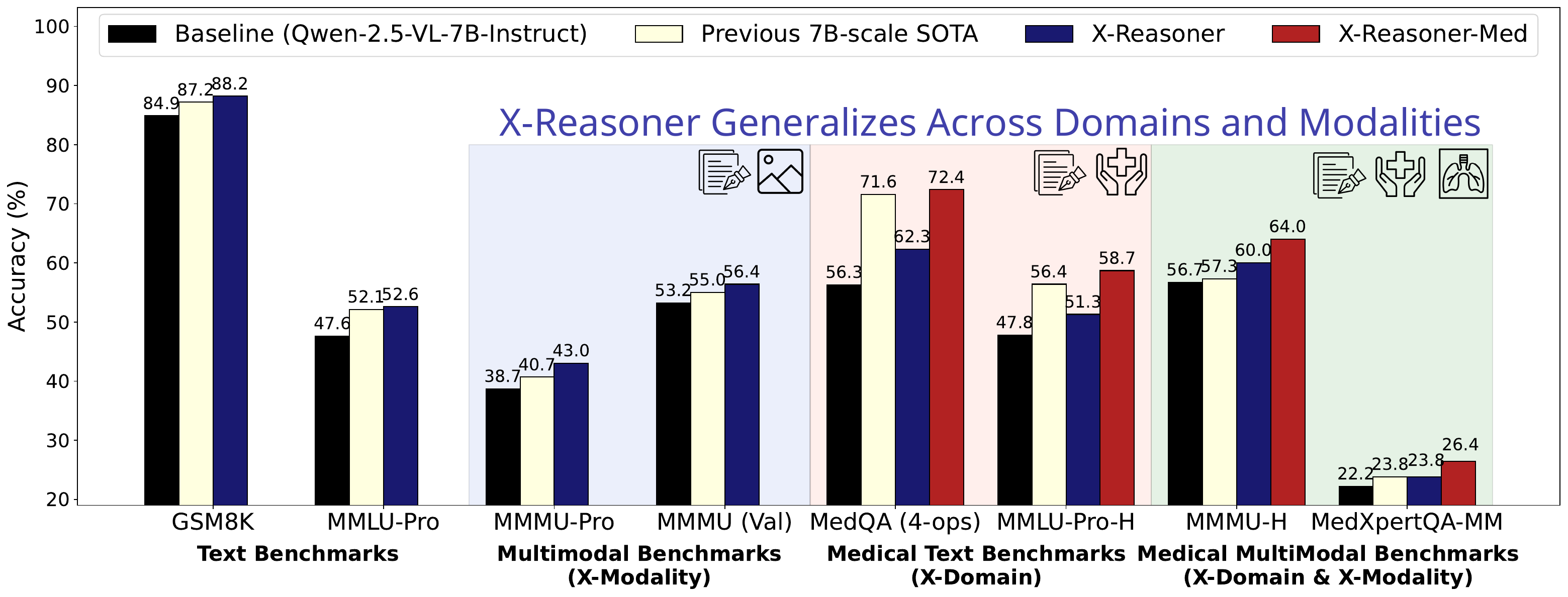}
    \vspace{-16pt}
    \caption{
    \ours (\bluesquare\ blue bars), fine-tuned solely on general domain text~\icon[1em]{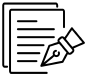}, shows strong generalization across both modalities (e.g., multimodality~\icon[1em]{figures/text-icon.pdf}\icon[1em]{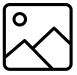}) and domains (e.g., medicine \icon[1em]{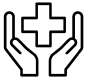}\icon[1em]{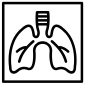}), surpassing prior SOTA (\Cref{tab:previous sota}) trained with in-domain multimodal data. \oursmed (\redsquare\ red bars), its medical-specialized variant, sets new SOTA on numerous medical benchmarks.
    }
    \label{fig:overview_figure}
        \vspace{-20pt}

\end{figure}

\end{abstract}

\newpage
\section{Introduction}

Reasoning has emerged as a foundational capability in language models, paving the way for a new paradigm known as test-time scaling~\citep{openai2024openaio1card}. This paradigm allows models to achieve consistent gains in complex reasoning tasks through scaling compute resources during inference. Recent proprietary models, such as o3~\citep{o3o4systemcard2025}, have begun to demonstrate strong multimodal reasoning capabilities. However, existing open-source research primarily focuses on advancing text-only reasoning, employing post-training techniques such as long CoT (Chain-of-Thought) distillation and reinforcement learning with verifiable rewards~\citep{guo2025deepseek, hu2025openreasonerzeroopensourceapproach}. While these efforts have resulted in powerful reasoning models, their evaluation has largely been confined to math and general-domain text benchmarks. Consequently, it remains uncertain how reasoning capabilities can effectively generalize beyond text inputs and general-domain contexts.

Recently, a growing body of open-source work has started addressing multimodal reasoning, i.e., extending reasoning to inputs that include both vision and language~\citep{wang2025vl,meng2025mm,guo2024mammoth}. Yet, current multimodal approaches often rely heavily on curating multimodal datasets which are tailored to specific tasks or domains, limiting their generalisability.

In this work, we ask a fundamental, yet under-explored question: \emph{Is reasoning generalizable across modalities and domains?} Specifically, we investigate whether such generalizable reasoning can be achieved through general-domain
text-based reasoning post-training. 
Beyond scientific significance, this question is also motivated by the practical advantages of general domain text-only training: i.e. its compute efficiency and the abundance and verifiability of general-domain textual reasoning data, which together allow us to avoid the cost and complexity of curating multimodal or domain-specific data. Our hypothesis is that text-based post-training, when carefully designed, can impart universal reasoning patterns
that robustly transfer across both unseen domains and input modalities. 

To investigate our hypothesis, we conduct an extensive empirical study using a two-stage text-only post-training recipe: supervised fine-tuning (SFT) on general-domain text data with distilled long CoTs, followed by reinforcement learning with verifiable rewards (RLVR) using mathematical textual questions. Remarkably, this pure textual training regimen proves sufficient to instill strong reasoning capabilities, enabling high performance not only on general-domain tasks but also on complex multimodal and domain-specific tasks. This finding suggests that the core structure of reasoning can indeed be acquired from general-domain text alone.

Building upon these insights, we introduce \textbf{\ours}, a 7B vision-language model post-trained with the proposed recipe.
Despite being trained only for text-based reasoning, \ours achieves superior performance on a suite of both text-only and multimodal reasoning benchmarks, outperforming prior state-of-the-art 7B models that were explicitly trained with multimodal reasoning supervision,
on challenging tasks including MMMU, MMMU-Pro and MathVista.
We further demonstrate that the improvement from \ours can generalize to specialized domains such as medicine. To boost in-domain performance even further, we explore the impact of incorporating domain-specific textual data.
To this end, we introduce \textbf{\oursmed}, a medical-specialized variant of \ours,  which undergoes additional post-training on medical domain text. \oursmed sets new records on numerous textual and multimodal medical tasks.

Our key contributions can be summarized as the following:
\begin{enumerate}[leftmargin=1.5em, itemsep=1pt, topsep=1pt]
\vspace{-1pt}
    \item \textbf{Generalizable Reasoning Study:}
    \begin{itemize}[leftmargin=1.5em, itemsep=1pt, topsep=1pt]
        \item We conduct an in-depth study of whether reasoning capabilities trained on general-domain text can generalize across input modalities and domains.
        \item We find that combining SFT and RL yields the strongest gains, and that math-based data serves as a strong anchor, though domain-specific data remains important for specialization.
    \end{itemize}
    
    \item \textbf{Introduction of \ours:}
    \begin{itemize}[leftmargin=1.5em, itemsep=1pt, topsep=1pt]
        \item We propose an effective post-training recipe for vision-language reasoning models that relies entirely on general-domain text-based data.
        \item \ours improves performance across modalities and domains, remarkably, achieving SOTA on challenging general-domain multimodal benchmarks. %
    \end{itemize}
    
    \item \textbf{Medical Domain Extension via \oursmed:}
    \begin{itemize}[leftmargin=1.5em, itemsep=1pt, topsep=1pt]
        \item We present \oursmed, a domain-adapted variant of \ours trained on medical text, setting new 7B-scale records across medical text-only and multimodal tasks.
    \end{itemize}
\end{enumerate}

\section{A Journey Towards Generalizable Reasoning}

\begin{figure}[!ht]
    \centering
    \includegraphics[width=1.0\textwidth]{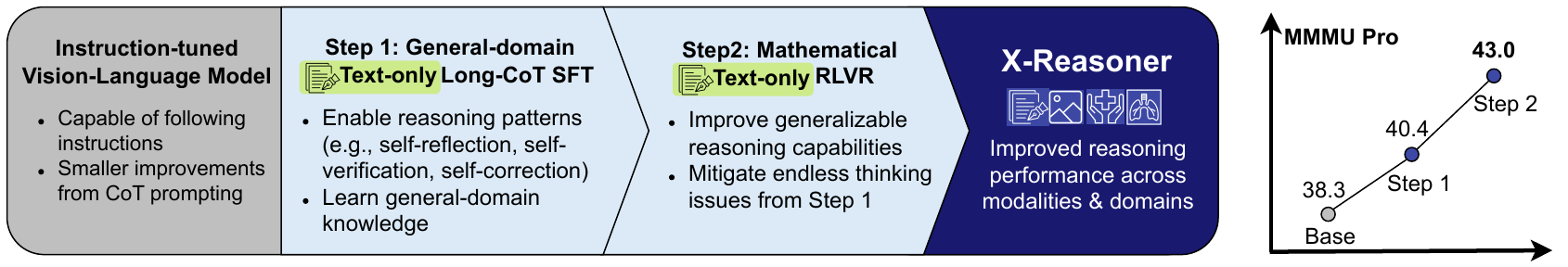}
    \caption{(Left) Our recipe for generalizable reasoning: based on an instruction-tuned VLM, we first conduct SFT on general-domain text data with distilled long CoTs. This is followed by RLVR on mathematical textual questions. This resulting model, \ours, exhibits significantly enhanced reasoning capabilities across modalities and domains. (Right) Model performance on MMMU-Pro (multimodal task) steadily improves at each stage of our recipe.}
    \label{fig:overview_recipe}
\end{figure}

In this section, we embark on a systematic exploration to understand whether 
generalizable reasoning can be effectively acquired through general-domain text-based post-training. 
Here, we define \emph{generalizable reasoning} as the ability to transfer reasoning capabilities not only across different task distributions but also across diverse modalities and specialized domains. To thoroughly address this question, we break it down into two sub-questions corresponding to two predominant post-training strategies used to foster reasoning: SFT and RL. Starting from an instruction-tuned vision-language model (VLM) capable of following instructions yet benefiting less from CoT prompting (see \Cref{sec:rq1}), we perform a detailed empirical study utilizing these two post-training methods. 
Our objective is to identify a robust and effective recipe that achieves strong, generalizable reasoning capabilities purely through general-domain text-based training.

\subsection{RQ1: Will Text-only SFT Improve Generalizable Reasoning?}
\label{sec:rq1}

Our first research question investigates whether reasoning trained solely via SFT on general-domain textual data can generalize beyond the training distribution. Specifically, we study two key aspects:
(1) the extent to which generalization can occur across domains and modalities, and (2) the essential role of reasoning in enabling such generalization.

\noindent\textbf{Extent of Generalization}
We begin by examining whether reasoning capabilities acquired through text-only SFT can generalize across distributional shifts and more importantly whether general-domain text-only SFT can transfer across domains and input modalities.

We start with a cross-distribution investigation where we fine-tune an instruction-tuned VLM, \base~\citep{baiQwen25VLTechnicalReport2025} using text-only MedQA \citep{jin2021disease}, a specialized medical QA dataset.
The training signal comprises detailed long-COT reasoning traces, distilled via rejection sampling \citep{huang2023large, zelikman2022star} from QwQ-32B~\citep{qwq32b}.
We then evaluate the model's generalization on an out-of-distribution task: MMLU-Pro-Health, the healthcare subset of MMLU-Pro \citep{wang2024mmlu}. 
The full training and evaluation details are provided in \Cref{sec:put-it-together,sec:eval-setup}.
As shown in the top rows of \Cref{tab:rq1}, all the SFT approaches consistently transfers the improvements from the source task (MedQA) to the target task (MMLU-Pro-Health), demonstrating strong cross-task generalization from text-only SFT. 

To further explore the scope of generalization, we perform SFT using \texttt{OpenThoughts-114k}~\citep{openthoughts}, a \emph{text-only} dataset covering math and science reasoning distilled from DeepSeek-R1 \citep{guo2025deepseek}. Results in \Cref{tab:rq1} show the SFT model yields consistent performance gains across three generalization axes: cross-domain (X-domain), cross-modality (X-modality) and the intersection of the two axes. This indicates that SFT with general-domain text-data already shows strong generalization capabilities. In addition, while CoT prompting alone provides some transferable gains, they are notably smaller compared with SFT, indicating that the VLM's inherent CoT capability prior to SFT is limited and CoT SFT is essential to fully unlock generalizable reasoning.

\noindent\textbf{Role of Reasoning in Generalization}
To isolate the contribution of reasoning, we compare models trained with and without explicit CoT supervision:
(1) a non-CoT SFT model trained directly on input-label pairs, and
(2) a CoT SFT model trained with reasoning traces.
Across all experiments in \Cref{tab:rq1}, the CoT-based SFT model consistently outperforms its non-CoT counterpart. 
Even among the baseline models, those equipped with CoT prompting outperform those without it, underscoring that explicit reasoning is a prerequisite for high performance on reasoning-centric tasks.
Furthermore, we observe that CoT-based models not only perform better in-distribution but also exhibit stronger transfer capabilities. Notably, the performance lift from CoT SFT over the CoT baseline is substantially larger than the corresponding improvement from non-CoT SFT over the non-CoT baseline across all settings. This suggests that explicit reasoning is the key to achieving both the highest in-distribution performance and the optimal generalization
across new domains and modalities.
 
\begin{table}[!ht]
\caption{Comparing baseline and text-only SFT with and without CoT in two settings: (1) Training on MedQA and evaluating on in \& out of distribution tasks; and (2) Training on general-domain OpenThoughts and evaluating on tasks across modalities and domains.}
\label{tab:rq1}
\tablestyle{5pt}{1}
\begin{tabular}{@{}cccrcr@{}}
\toprule
\multirow{2}{*}[-.3em]{\shortstack{\em Training: MedQA \\ \em Eval: in \& out of distribution}} &  & \multicolumn{2}{c}{Non-CoT} & \multicolumn{2}{c}{CoT} \\ \cmidrule(l){3-6} 
 & Modality & Baseline & \multicolumn{1}{c}{SFT} & Baseline & \multicolumn{1}{c}{SFT}  \\ \midrule
MedQA (In distribution) & Textual & 49.3 & 57.6 (+8.3) & 50.5 & \textbf{61.9} (\textbf{+11.4})  \\
MMLU-Pro-H (Out of dist.) & Textual & 43.7 & 47.2 (+3.5) & 47.8 & \textbf{54.2} \textcolor{white}{0}(\textbf{+6.4}) \\ \midrule\midrule
\multirow{2}{*}[-.3em]{\shortstack{\em Training: OpenThoughts (general-domain)
\\ \em Eval: across tasks, modalities and domains}} &  & \multicolumn{2}{c}{Non-CoT} & \multicolumn{2}{c}{CoT} \\ \cmidrule(l){3-6} 
 & Modality & Baseline & \multicolumn{1}{c}{SFT} & Baseline & \multicolumn{1}{c}{SFT}\\ \midrule
MMLU-Pro (general-domain text) & Textual & 39.5 & 39.5 (+0.0) & 47.6 & \textbf{50.4} \textcolor{white}{0}(\textbf{+2.8})  \\
MedQA (X-domain) & Textual & 49.3 & 52.1 (+2.8) & 50.5 & \textbf{55.2} \textcolor{white}{0}(\textbf{+4.7}) \\
MMMU-Pro (X-modality) & Multimodal & 34.6 & 36.0 (+1.4) & 38.3 & \textbf{40.4} \textcolor{white}{0}(\textbf{+2.1}) \\
MMMU-Pro-H (X-modality\&domain) & Multimodal & 28.1 & 30.2 (+2.1) & 34.3 & \textbf{37.9} \textcolor{white}{0}(\textbf{+3.6})  \\ \bottomrule
\end{tabular}
\end{table}

\begin{AIbox}{Takeaway 2.1.1}
SFT on general-domain text-only data, when enriched with long CoTs, can endow models with generalizable reasoning capabilities that transfer across tasks, domains, and modalities.
\end{AIbox}

\noindent\textbf{Forced-Exiting: Mitigating Endless Thinking in Long-CoT SFT \label{par: force stopping}}
While long-CoT SFT effectively learns transferrable reasoning patterns, such as self-reflection, verification, and correction (see qualitative analyses in \Cref{sec:examples}), we observe a recurring challenge: the model occasionally fails to terminate its responses. Specifically, the model continues generating tokens indefinitely without producing a stop signal. Such endless thinking occurs approximately 17\% of generations for both MedQA and MMLU-Pro-Health tasks, as reported in (\Cref{tab:nonrepetition}). Upon closer inspection, these failure cases are often characterized by verbose, repetitive output, where the model persistently attempts to ``re-think'' or ``double-check'', an artifact likely stemming from the reasoning mechanism introduced during long-CoT SFT~\citep{guo2025deepseek}.

To address this, we implement a forced-exiting mechanism inspired by the method introduced in \citet{s125}. This involves appending a designated stop token `\verb|</think>|' once the output sequence reaches a predefined length threshold. This soft intervention encourages the model to wrap up its reasoning process within a reasonable token budget.
As shown in \Cref{tab:nonrepetition}, applying forced-exiting effectively reduces endless thinking, thereby improving final task accuracy. 
\vspace{-1em}
\begin{table}[h]
\centering
\tablestyle{4pt}{1}
\caption{Forced-exiting mitigates endless thinking and improves final task accuracy from CoT SFT}
\label{tab:nonrepetition}
\begin{tabular}{lcccc}
\toprule
Task & \multicolumn{2}{c}{Before Forced-Exiting} & \multicolumn{2}{c}{After Forced-Exiting}\\
\cmidrule(l){2-3} 
\cmidrule(l){4-5}
 & \% endless thinking & Accuracy & \% endless thinking & Accuracy \\
\midrule
MedQA & 17.1\% & 56.0 & 0\% & 61.9\\
MMLU-Pro-Health & 17.3\% & 49.0 & 0.1\% & 54.2 \\
\bottomrule
\end{tabular}
\end{table}

\vspace{-1em}
\begin{AIbox}{Takeaway 2.1.2}
Forced-exiting provides a practical and effective solution to address the endless thinking behavior in long-CoT SFT, improving both reliability and downstream task performance.
\end{AIbox}

\subsection{RQ2: Will Mathematical Text-only RL Improve Generalizable Reasoning?}

Reinforcement Learning (RL) has become an effective approach for training models to reason. In this section, we explore whether mathematical text-only RL can promote generalizable reasoning.

\noindent\textbf{Learning Algorithm} We employ GRPO~\citep{grpo24} as our RL algorithm~\citep{rl88}. GRPO has recently become favored over algorithms such as PPO~\citep{ppo17} as it does not require an additional value function. Instead, GRPO computes the advantage by comparing responses within each query-specific group. The detailed formulation is provided in \Cref{sec:grpo}.

 To stabilize and improve GRPO training, we incorporate several enhancements inspired by recent work~\citep{drgrpo25,dapo25}: (1) using a higher upper clipping threshold (clip-higher), which promotes sample diversity and mitigates entropy collapse; (2) employing a token-level policy gradient loss, which normalizes the loss by the total number of tokens within a batch, thus addressing the response-length bias present in GRPO; and (3) reducing or entirely removing the KL penalty, a change beneficial in training models performing long chain-of-thought (CoT) reasoning, where the policy distribution may significantly diverge from the reference model.

\noindent\textbf{Verifiable Rewards}
Rather than relying on a learned reward model, which often encounters reward-hacking issues~\citep{gao2022scalinglawsrewardmodel}, we directly utilize the final accuracy on verifiable tasks as our reward metric. Specifically, a response receives a reward of $r = 1$  if it matches the ground truth answer under semantic equivalence (i.e., \texttt{is\_equivalent}$(\hat{y}, y)$), and $r = 0$ otherwise.
We omit format rewards, as our policy model already adheres well to instruction formatting.

\noindent\textbf{The Role of Training Data: Math as a Generalization Anchor}
\label{par: domain RL}
A central question in text-only RL is: What type of training data best supports reasoning generalization? While reasoning models are often fine-tuned on domain-specific datasets, we hypothesize that  mathematics serves as an especially effective domain for promoting generalizable reasoning, as math tasks naturally elicit long, structured chains of thought that may transfer more effectively across domains.

To test this, we compare RL fine-tuning on two data sources: (1) math textual questions from Orz~\citep{hu2025openreasonerzeroopensourceapproach}, and (2) MedQA, a domain-specific medical textual QA dataset. 
Results in \Cref{tab:cross_domain_rl} reveal a clear asymmetry: models trained on math data not only improve on math benchmarks (e.g., GSM8K, MathVista), but also show strong gains on out-of-domain medical tasks. In contrast, models trained on MedQA excel in medical benchmarks but do not consistently improve math tasks.
This suggests that \textbf{math serves as an “anchor domain”}, whose reasoning structure is sufficiently rich and general to support cross-domain\&modality transfer.
Supporting this, we observe in \Cref{tab:cross_domain_rl} that math-trained models consistently generate longer responses across benchmarks.
Furthermore, across all models, responses to math tasks are longer than those to medical tasks, highlighting math’s inherent demand for extended reasoning.
That said, domain-specific data remains important. RL fine-tuning on MedQA yields the best performance on medical tasks, indicating that in-domain RL is still critical for achieving optimal performance in specialized domains.

\vspace{-1em}
\begin{table}[htbp]
\centering
\tablestyle{1pt}{1}
\caption{Comparing the effect of the training data domain (math vs medicine) on RL generalization. We report accuracy [response length in number of words] for each experiment.}
\label{tab:cross_domain_rl}
\begin{tabular}{lcccc}
\toprule
\multirow{2}{*}{Method}& MMLU-Pro-Health & OmniMedVQA & GSM8K & MathVision\\
 & (medical text) & (medical multimodal) & (math text) & (math multimodal)\\
\midrule
Baseline & 47.8 [105] & 57.4 \textcolor{white}{0}[84] & 86.0 [146] & 24.7 [293] \\
+ text-only RL Math (Orz) & 50.7 [166] & 58.4 [112]& {\bf 88.8} [200] & {\bf 28.1} [580]\\
+ text-only RL Medicine (MedQA)  & {\bf 53.6} \textcolor{white}{0}[87] & {\bf 59.2} \textcolor{white}{0}[72] & 85.7 [135]& 26.2 [293]\\
\bottomrule
\end{tabular}
\end{table}

\begin{AIbox}{Takeaway 2.2.1}
Mathematics provides a strong foundation for generalizable reasoning in RL, serving as an anchor domain. However, domain-specific data remains essential for optimal performance in specialized contexts.
\end{AIbox}

\noindent\textbf{RL vs SFT}
Prior work has positioned RL as a powerful yet volatile method for promoting generalizable reasoning~\citep{chu2025sft}, while SFT is widely recognized for its stability and efficacy in capturing structured reasoning patterns. However, RL is often plagued by training instability and convergence issues, particularly when applied in isolation~\citep{yeo2502demystifying}.
To better understand their relative strengths,  we conduct a direct comparison between RL and SFT in fostering generalization across task distributional shifts, domain shifts and modality transitions. Specifically, we compare general-domain text-only SFT (trained on general-domain OpenThoughts data) and RL (trained on math questions from Orz), and their combination (SFT + RL). 
Alongside general-domain text task (MMLU-Pro), we evaluated the models in three distinct generalization settings: cross-domain (medical text: MMLU-Pro-Health), cross-modality (multimodal: MMMU-Pro), cross domain\&modality (multimodal medical: NEJM Image Challenge). 
In \Cref{tab:rl vs sft general}, we first observe that general-domain SFT and RL can both significantly improve from baseline across all the generalization settings, indicating both SFT and RL can elicit generalizable reasoning.  We further notice that pure RL is overall slightly worse than SFT.
This performance gap can be attributed to RL’s difficulty in learning long-CoT reasoning from scratch, evidenced by its shorter, less structured outputs. In contrast, SFT excels at acquiring rich CoT patterns through supervised distillation, yielding significantly longer and more coherent responses.
Interestingly, when RL is applied after SFT, it benefits from the structured reasoning foundation established by SFT, and further refines the model’s capabilities via RLVR. 
We also observe that RL + SFT is able to regulate the endless thinking issue from SFT as the percentage of responses that exceeds the maximum length decreases during RL training, as shown in the training response clip ratio in \Cref{fig:dynamics}.
In summary, the hybrid approach, general-domain text-only SFT + RL, achieves the best results across all generalization settings, effectively combining SFT’s stability and inductive strength with RL’s reward-guided optimization\footnote{We also conduct a more controlled comparison between SFT and RL, both trained on the same MedQA data, and arrive at the same conclusion (\Cref{tab:rl vs sft}).}.

\vspace{-1em}
\begin{table}[htbp]
\centering
\tablestyle{3pt}{1}
\caption{Comparison of performance from general-domain text-based RL, SFT, and SFT + RL in general-domain textual task (MMLU-Pro) and generalization setups including cross-modality (X-modality), cross-domain (X-domain) and combined cross-domain\&modality (X-modality\&domain) tasks. SFT is trained with OpenThoughts data and RL is trained with math data.}
\label{tab:rl vs sft general}
\begin{tabular}{lccccc}
\toprule
\multirow{2}{*}{Method} & MMLU-Pro & MMLU-Pro-Health  & MMMU-Pro & NEJM Image Challenge & Response Len \\
& (G-domain text) & (X-domain) & (X-modality) & (X-domain\&modality) & (\#words) \\
\midrule
Baseline & 47.6 & 47.8 & 38.3 & 41.8 & 122\\ %
SFT & 50.4 & 50.4 & 40.4 & 45.0 & 946\\ %
RL & 50.3 & 50.3 & 40.6 & 44.4 & 232 \\ %
SFT + RL & {\bf 53.3} & {\bf 53.1}  & {\bf 42.5} & {\bf 45.7} & 977 \\ %
\bottomrule
\end{tabular}
\end{table}

\vspace{-1em}
\begin{AIbox}{Takeaway 2.2.2}
Combining SFT with RL, both trained on general-domain text-only data, proves to be the most effective strategy for achieving optimal performance and robust generalization.
\end{AIbox}

\subsection{Putting it all together: \ours}
\label{sec:put-it-together}
Our investigation reveals that \emph{general-domain text-only post-training, when carefully designed, can drive strong generalizable reasoning across tasks, domains, and modalities.}
Therefore, we conclude our investigation by consolidating our findings into a coherent training recipe, illustrated in \Cref{fig:overview_recipe}, culminating in a powerful generalizable reasoning model named \textbf{\ours}.
Specifically,
initialized with \base~\citep{baiQwen25VLTechnicalReport2025}, our training recipe follows:

\noindent\textbf{Step 1: text-only general-domain Long-CoT SFT}.
We begin by applying SFT to elicit explicit structured reasoning capabilities using long-CoT reasoning traces.
We train our model on the general-domain \texttt{open-thoughts/OpenThoughts-114k} dataset~\citep{openthoughts}. This dataset contains curated reasoning traces on 114k math, coding and science questions, distilled by the DeepSeek-R1 model. 
We fine-tune the model for 4 epochs with a learning rate of $1\times 10^{-5}$.
i%

\noindent\textbf{Step 2: text-only math RLVR}.
After SFT, we further refine our model using RL with verifiable rewards to enhance its reasoning accuracy and generalization.
We trainin our model on \texttt{Orz-math-57k}, a set of 57k mathematical textual questions curated by~\citet{hu2025openreasonerzeroopensourceapproach}.
During this stage, we set the total training episodes/epochs to 3, use a learning rate of $3\times 10^{-6}$, a global batch size of 128, and sample 8 rollouts per query with a maximum response length of 4,096 tokens.

Detailed hyperparameters for each training stage are provided in \Cref{sec:hyperparameters}.
The SFT stage was conducted using 8 40GB A100 GPUs over 8 hours, and the RLVR stage utilized 32 40GB A100 GPUs for approximately 56 hours. We provide visualization of training dynamics in \Cref{sec:dynamics}.

\section{A Comprehensive Evaluation of \ours}

\noindent\textbf{Evaluation Setup}
\label{sec:eval-setup}
To thoroughly assess the generalizability of \ours, we evaluate its performance across four distinct settings: (1) General-domain text-only tasks; (2) General-domain multimodal tasks (X-modality); (3) Specialized-domain text-only tasks (X-domain); and (4) Specialized-domain multimodal tasks (X-modality\&domain).
Detailed descriptions of these tasks can be found in \Cref{sec:tasks}.
For inference, we utilize the vLLM backend~\citep{kwon2023efficient}. Consistent with recent findings on reproducibility~\citep{hochlehnert2025sober}, we observe performance variability even under greedy decoding (i.e., temperature=0), attributed to hardware differences, software stack inconsistencies, runtime environments, etc.
In this study, we report greedy decoding results to compare with previous studies. In addition,
to ensure the robustness and reproducibility,  we report three additional metrics computed over five runs at temperature = 0.3: average accuracy, majority-vote accuracy, and pass@n accuracy. Majority-vote accuracy aligns with the self-consistency method introduced by \citet{selfconsistency23}, which has proven effective for scaling up parallel test-time evaluation.
Pass@n accuracy provides further insight into the model’s effective exploration of the answer space, indicating potential performance gains achievable through additional reinforcement learning.
Unless otherwise indicated, we consistently use CoT prompting templates for both baselines and \ours evaluations.
Complete exact prompt templates are provided in \Cref{sec:prompts}.
Additionally, we apply a forced-exiting mechanism as described by \citet{s125}, capping output generation at 4,096 tokens. This mechanism not only improves memory efficiency but also effectively handles endless thinking cases, as elaborated in \Cref{par: force stopping}.

\vspace{-1em}
\begin{figure}[!ht]
    \centering
    \includegraphics[width=1.0\textwidth]{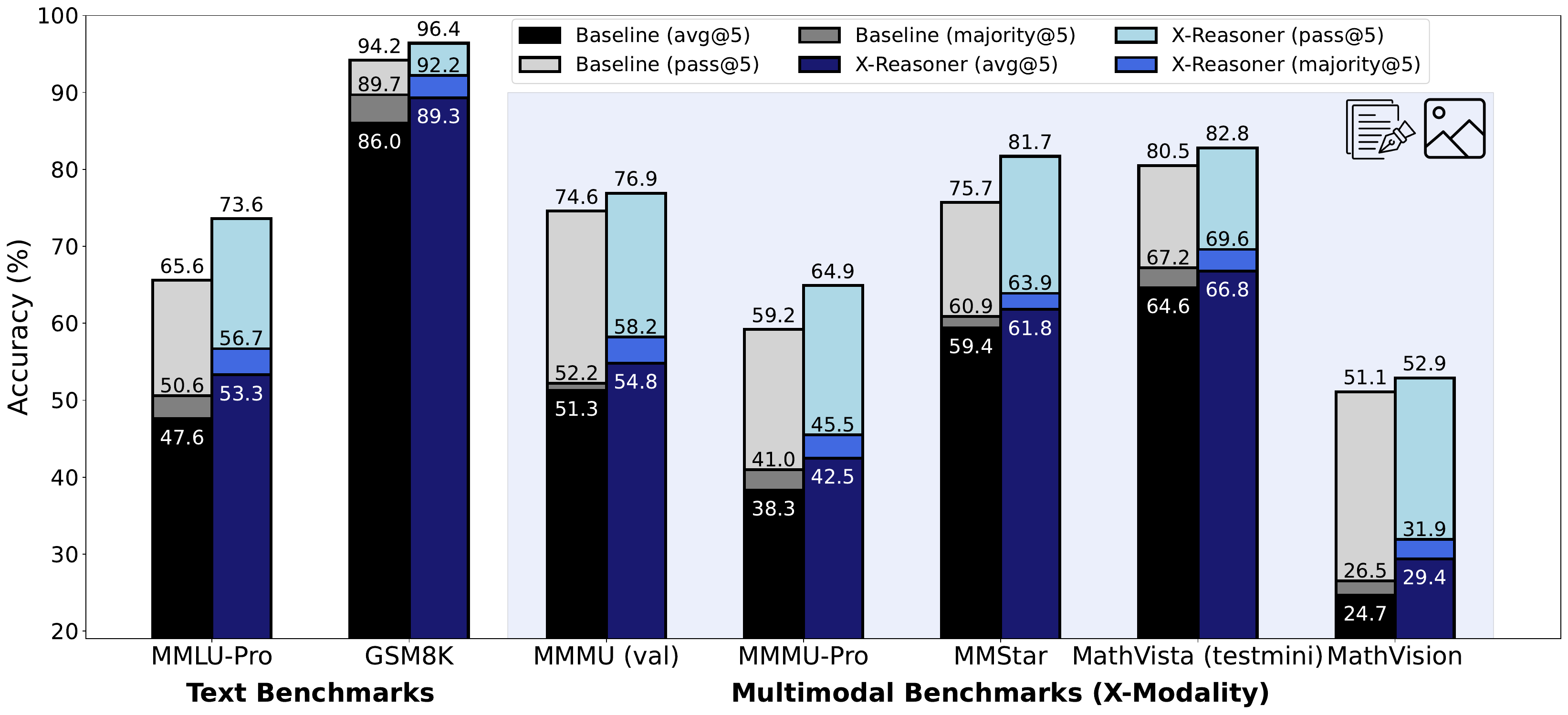}
    \caption{Comparing \ours and baseline on text-only benchmarks and multi-modal benchmarks. Despite being trained with general-domain text-only data, \ours can significantly improve multi-modal benchmarks, showing the generalization of \ours's reasoning.}
    \label{fig:general_domain_main}
\end{figure}

\vspace{-1em}
\subsection{\ours's Cross-Modality Generalization \label{sec:cross modality}}

In \Cref{fig:general_domain_main}, we assess the cross-modality generalization capabilities of \ours. We first validate its effectiveness on text-only tasks, observing significant performance gains across benchmarks such as MMLU-Pro and GSM8K. Crucially, these improvements robustly transfer to multimodal contexts, reflected consistently across average, majority vote, and pass@5 accuracy. Such consistent improvements indicate not only overall superior reasoning capability, but also suggest that \ours has a more effective and expansive search space for further improvements.

Comparisons against SOTA multimodal reasoning models are presented in \Cref{tab:compare_sota}. \ours achieves SOTA on MMMU, MMMU-Pro and MathVista, while remaining competitive on others. Notably, these SOTA models are extensively trained on multimodal data, underscoring the critical finding that 
text-only reasoning training alone is sufficient—and likely accounts for the majority of learning necessary for effective multimodal reasoning.

Qualitative analysis in \Cref{sec:examples} highlights that \ours consistently generates more sophisticated reasoning outputs compared to the baseline. Reasoning patterns demonstrated by \ours include detailed planning, verification (e.g., "wait" steps), and error correction. In the cross-modality setting, \ours seamlessly integrates visual information into its reasoning chains, providing step-by-step reasoning with verification. For example, in \Cref{tab:x-modality-example}, \ours methodically reviews all bar values in a plot, avoiding the oversight errors prevalent in the baseline. \ours also identifies and corrects mistakes within its reasoning process. 
These observations reinforce our conclusion that \ours successfully internalizes generic reasoning patterns through text-only training, effectively extending them across modalities.

\vspace{-1em}
\begin{table}[!ht]
\tablestyle{3pt}{1}
\caption{\ours outperforms SOTA multimodal models on the 7B/8B scale on challenging multimodal tasks. All results are based on greedy decoding.}
\centering
\begin{tabular}{llcccc}
\toprule
\multirow{2}{*}{\textbf{Model}} & \textbf{Reasoning}  & MMMU & MMMU-Pro & MathVista & MathVision\\
& {\bf Data Source} & (Val) & & (testmini)\\
\midrule
Qwen-VL-2.5-7B-Instruct~\citep{baiQwen25VLTechnicalReport2025} & -  & 53.0 & 38.7 & 62.8 & 25.0 \\
R1-Onevision-7B \citep{yang2025r1} & multimodal & - & - & 64.1 & 29.9\\
MAmmoTH-VL2-7B \citep{jia2025visualwebinstruct} & multimodal & 54.7 & 40.7 & 68.1 & -\\
MM-Eureka-8B \citep{meng2025mm} & multimodal & 49.2 & - & 67.1 & 22.2 \\
Mulberry-7B \citep{yao2024mulberry} & multimodal  & 55.0 & 36.8 & 63.1 & - \\
\midrule

\ours (7B) & text & {\bf 56.4} & {\bf 43.0} & {\bf 69.0} & 29.6 \\
\bottomrule
\label{tab:compare_sota}
\end{tabular}
\end{table}

\vspace{-1em}
\begin{AIbox}{Takeaway 3.2.1}
\ours, trained on text-only data, consistently improves multimodal task performance, matching or surpassing prior SOTA trained explicitly with multimodal data.
\end{AIbox}

\noindent\textbf{Ablation Study: Is \ours simply solving the text-solvable examples?}
A potential limitation in evaluating vision-language models is their tendency to rely on text-based shortcuts, solving tasks without genuinely integrating visual information.
To rigorously confirm that the cross-modality generalization observed with \ours is not merely due to improved text-only shortcut solutions, we conduct an ablation experiment. Specifically, we identify and remove text-solvable examples and assess whether performance gains persist afterward.

To identify these text-solvable examples, we mask visual inputs for multimodal tasks previously evaluated in \Cref{sec:cross modality}, performing text-only evaluations using both \ours and the baseline \base. For each model, we sample three independent responses per question and eliminate examples consistently solved through text alone.

\Cref{tab:remove_text_cue} summarizes the number of remaining examples after removing text-solvable examples for each task, along with the corresponding model performances.
While a substantial number of text-solvable instances were identified, these primarily reflect models' textual reasoning shortcuts rather than dataset issues~\citep{yue2024mmmupro}.
Crucially, \ours maintains performance advantages over the baseline after text-only solvable examples are excluded, affirming true multimodal reasoning capabilities. This confirms that \ours effectively utilizes visual context within the reasoning framework acquired through text-only training.

\vspace{-1em}
\begin{table}[!ht]
\centering
\tablestyle{6pt}{1}
\caption{Comparing \ours and baseline performance after removing text-solvable examples. The improvement of \ours remains, indicating true multimodal generalizability.}
\label{tab:remove_text_cue}
\begin{tabular}{llcc}
\toprule
\textbf{Task} & \textbf{\#original - \#text-solvable = \# } & \textbf{Baseline} & \textbf{\ours}\\
\midrule
MMMU &900 - 308 = 592 & 39.3 & {\bf 41.4} \\
MMMU-Pro & 1730 - 259 = 1471 & 33.4 & {\bf 36.4} \\
MathVista (testmini) &1000 - 257 = 743 & 57.9 & {\bf 60.6} \\
MathVision &3040 - 448 = 2592 & 18.6 & {\bf 21.8} \\
MMStar & 1500 - 259 = 1241 & 56.6 & {\bf 59.0} \\
\bottomrule
\end{tabular}
\end{table}

\vspace{-1em}
\begin{AIbox}{Takeaway 3.2.2}
\ours achieves authentic multimodal reasoning capabilities, as evidenced by sustained improvements after excluding text-only solvable examples.
\end{AIbox}

\subsection{\ours in Medicine: A Cross-Modality and Cross-Domain Study}

In this section, we examine the cross-domain and cross-modality transfer capabilities of \ours, specifically within medical contexts.
While \ours is trained exclusively on general-domain text data, we further probe the benefits of domain-specific adaptation by developing \oursmed, which extends \ours with additional training using medical text data. 

\noindent\textbf{Evaluation Setup}
Our evaluation includes three text-only medical tasks: MMLU-Pro-Health (the healthcare subset of MMLU-Pro \citep{wang2024mmlu}, selected following the setup in \citep{chen2024huatuogpt}), MedQA (the original USMLE version), and MedQA (4-ops: the four-choice version).
For multimodal medical evaluations, we utilize healthcare subsets of MMMU \citep{yue2024mmmu} and MMMU-Pro \citep{yue2024mmmupro}, designated as MMMU-Health and MMMU-Pro-Health respectively. We also include MedXpertQA-MM \citep{zuo2025medxpertqa}, OmniMedVQA
\citep{hu2024omnimedvqa} and NEJM Image Challenge \citep{nejm_image_challenge}, which are specifically designed to evaluate medical knowledge and reasoning. 

\vspace{-1em}
\begin{figure}[!ht]
    \centering
    \includegraphics[width=1.0\textwidth]{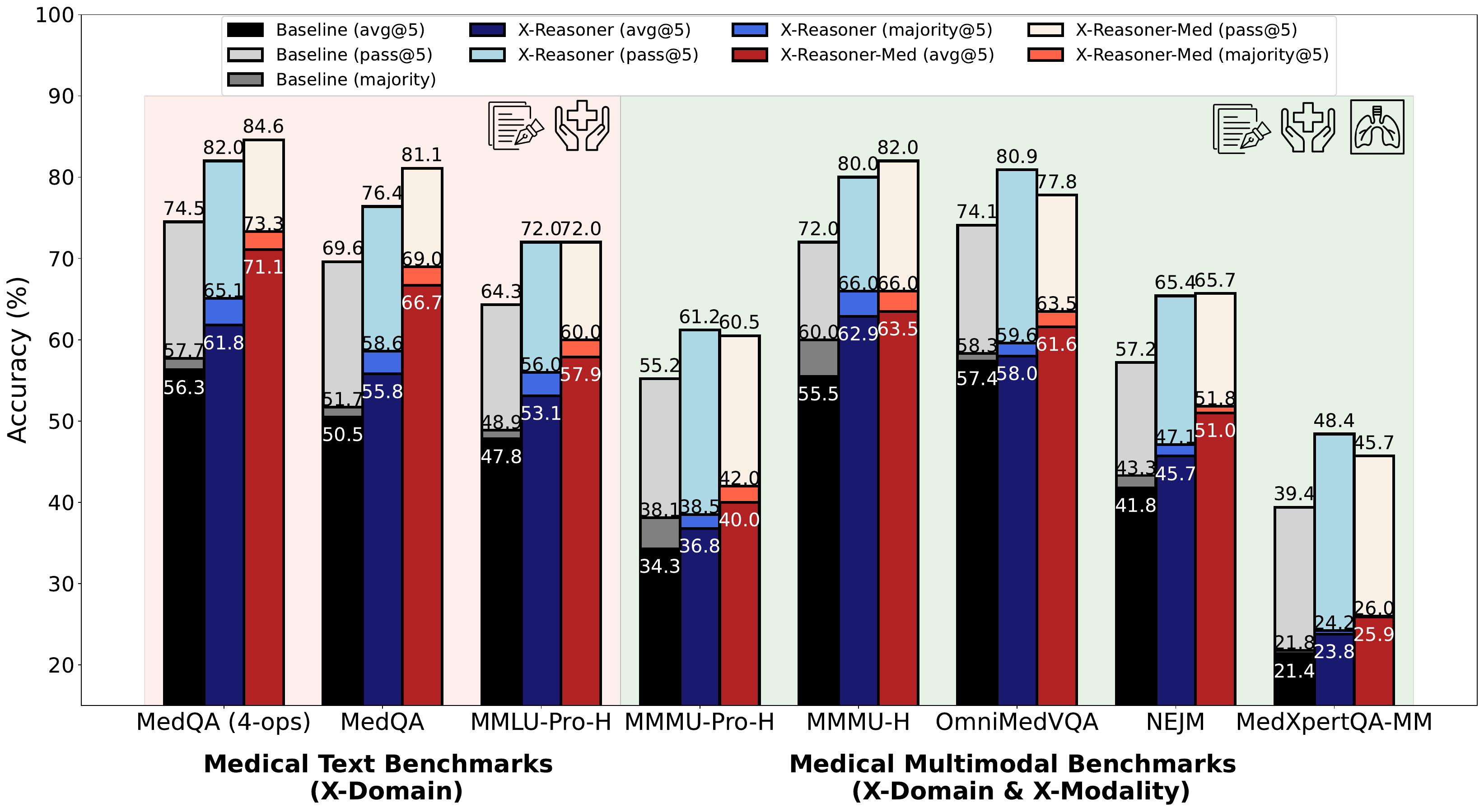}
    \caption{Comparing \ours, \oursmed and baseline on text and multimodal medical benchmarks. \ours, trained with general-domain text-only data brings consistent improvement across medical tasks. \oursmed, obtained by continued training of \ours on medical text data, further improves medical domain performance.}
    \label{fig:medical_domain_main}
\end{figure}

\noindent\textbf{Results}
\Cref{fig:medical_domain_main} shows \ours demonstrates robust cross-domain and cross-modality generalization capabilities, surpassing baseline models on both textual and multimodal medical tasks. Remarkably, as detailed in \Cref{fig:overview_figure}, \ours surpassing or matching previous SOTA on MedXpertQA-MM and MMMU-Health, despite previous SOTA models being explicitly trained on multimodal medical data. Qualitative assessments in \Cref{sec:examples} further illustrate \textsc{X-Reasoner}'s effective integration of visual information and medical domain-specific reasoning. As an example, \ours accurately identifies medical patterns, such as correctly recognizing white matter abnormalities indicative of progressive multifocal leukoencephalopathy in histology slides (\Cref{tab:x-modality-domain-example}).

\begin{AIbox}{Takeaway 3.3.1}
\ours substantially enhances medical task performance, confirming its strong capability for cross-domain and cross-modality generalization.
\end{AIbox}

\noindent\textbf{\oursmed: Enhancing Medical Performance}
While \ours shows substantial improvements on medical tasks, our analysis in \Cref{par: domain RL} highlights the critical role of domain-specific data for achieving optimal in-domain performance. This leads us to investigate whether additional medical-domain text data can further enhance \textsc{X-Reasoner}’s performance. To explore this, we extend \textsc{X-Reasoner}'s training with text-only MedQA data. Specifically, we perform SFT with distilled CoTs from QwQ-32B~\citep{qwq32b}, followed by RLVR using the same MedQA data.

The resulting model, \oursmed, achieves further improvements across medical benchmarks. As depicted in \Cref{fig:medical_domain_main}, \oursmed consistently surpasses \ours in terms of average and majority vote accuracies. 
Notably, \ours sometimes attains higher pass@n scores, suggesting it explores a broader search space. Conversely, \oursmed, benefiting from targeted medical-domain fine-tuning, already leverages this search space more effectively but potentially with reduced room for further gains. Nevertheless, as highlighted in \Cref{fig:overview_figure}, \oursmed sets new SOTA performance for both text-only and multimodal medical tasks, validating our hypothesis that combining general text-based reasoning with domain-specific text-based fine-tuning unlocks substantial additional performance gains for specialized domains.

\begin{AIbox}{Takeaway 3.3.2}
\oursmed, resulting from continued training of \ours with medical text-only data, achieves further performance enhancements and establishes new SOTA results in text and multimodal medical benchmarks.
\end{AIbox}

\section{Related Work}

In the context of RL, several methods have been proposed for learning multimodal reasoning.
Notable examples include VFT~\citep{liu2025visual}, MM-Eureka~\citep{meng2025mm}, Vision-R1~\citep{huang2025vision}, R1-V~\citep{chen2025r1v}, and the framework proposed by \citet{zhou2025r1}.
These works demonstrate how RL can be used to refine reasoning abilities within VLMs. 
\citet{peng2025lmm} present a two-stage RL approach that leverages both text-based and multimodal data, while \citet{deng2025openvlthinker} explore an iterative training scheme alternating between SFT and RL on multimodal inputs, beginning with distillation of CoT reasoning from image captions.

On the SFT side, reasoning is commonly distilled from captions or responses synthesized by VLMs. Methods such as MaMMOTH~\citep{guo2024mammoth,jia2025visualwebinstruct}, Vision-R1~\citep{huang2025vision}, R1-OneVision~\citep{yang2025r1}, and Llava-CoT~\citep{xu2024llava} follow this strategy to infuse reasoning into multimodal models during SFT. An alternative approach involves linking a text-based reasoning model with a vision encoder to incorporate visual grounding~\citep{skywork2025r1v}. 

Recently, there has been growing interest in applying reasoning techniques to the medical domain, initially focusing on the text modality~\citep{zhang2025med, huang2025m1unleashpotentialtesttime, chen2024huatuogpt}. Subsequently, researchers have begun exploring RL-based training on multimodal medical datasets to enhance multimodal reasoning performance, e.g., \citet{pan2025medvlm} and \citet{lai2025med}.

\section{Discussions and Limitations}
\label{sec:discussion}

While we show \ours 
effectively equips vision-language models with generalizable reasoning capabilities, we acknowledge several limitations that offer opportunities for future work:

\noindent\textbf{Base Model Constraints}
Prior studies have emphasized the importance of starting from pretrained base checkpoints, rather than instruction-tuned models, to better incentivize the emergence of reasoning behaviors~\citep{guo2025deepseek, hu2025openreasonerzeroopensourceapproach}. However, in our case, we are constrained by computational resources and the lack of base VLMs in the recent open-source releases, limiting our ability to empirically validate these claims.

\noindent\textbf{Model Scale and Backbone Diversity}
Prior work suggests that larger models tend to exhibit stronger reasoning capabilities~\citep{guo2025deepseek}. However, due to computational constraints, our model size is limited to 7B parameters. We also focused exclusively on the Qwen-VL series and did not evaluate other publicly available VLMs.

\noindent\textbf{Task Scope}
Our evaluation primarily targets mathematical questions and multiple-choice questions across general and medical domains. While these benchmarks provide clear, verifiable supervision for assessing reasoning capabilities, they do not capture the full spectrum of real-world reasoning challenges. In particular, we have not tested our approach on open-ended generation, interactive dialogue, or instruction-following scenarios. Understanding how reasoning generalizes to such unconstrained settings remains an important area for future exploration.

\section{Conclusion}

In this work, we introduce \ours, a simple yet effective post-training recipe that enhances the reasoning capabilities of VLMs using only general-domain text-based supervision from a combined SFT + RL strategy. 
Through extensive experiments, we demonstrate that the reasoning skills acquired via \textsc{X-Reasoner}'s text-only training pipeline generalize effectively across modalities and domains, surpassing prior SOTA trained with in-domain multimodal data. We further introduce \oursmed, a domain-specialized variant of \ours trained on medical text which achieves new SOTA on a range of medical benchmarks across both text-only and multimodal settings.
Our findings underscore the surprising power of textual supervision for learning transferable reasoning in multimodal models. We leave to future work the exploration of continued in-domain multimodal SFT/RL to further strengthen domain-specific and multimodal reasoning capabilities.

\bibliography{iclr2025_conference}
\bibliographystyle{iclr2025_conference}

\appendix
\newpage
\section{GRPO}
\label{sec:grpo}

Let $\pi_\theta$ be a vision language model (VLM) based on decoder-only transformers and parameterized by $\theta$.
It takes as input a sequence of tokens, denoted by $\vec{q}$, such as a question, and autoregressively decodes the response, denoted by $\vec{o}$.
Optionally, $\pi_\theta$ can also take visual features, such as pictures, as input that is tokenized by a vision encoder.
However, visual features are not included in our training paradigm, and $\vec{q}$ during training is supposed to be natural languages only.
We use \text{text-only} math datasets for training.
Let $\mathcal{D}$ be a dataset and $(\vec{q}, \vec{a}) \in \mathcal{D}$ be pairs of questions and answers.

In Group Relative Policy Optimization~\citep[GRPO]{grpo24}, for a given query $\vec{q}$, the policy model samples a group of responses, and the token-level advantage $\hat{A}_{i,t}$ for the $i$-th response is estimated by normalizing its reward $r_i$ relative to the group rewards $\vec{r}$:
\begin{align}
    \hat{A}_{i,t}=\frac{r_i-\text{mean}(\vec{r})}{\text{std}(\vec{r})},
\end{align}

Similar to PPO~\citep{ppo17}, GRPO employs a clipped surrogate objective but includes a KL-divergence penalty term directly in its loss function to encourage stability:

\newcommand{\oldpi}{\pi_{\theta_\text{old}}}
\newcommand{\piratio}{\frac{\pi_\theta(o_{i,t}\mid \vec{q}, \vec{o}_{i,<t})}{\oldpi(o_{i,t}\mid \vec{q}, \vec{o}_{i,<t})}\hat{A}_{i,t}}
\newcommand{\refratio}{\frac{\pi_\text{ref} (o_{i,t}\mid \vec{q}, \vec{o}_{i,<t})}{\pi_\theta(o_{i,t}\mid \vec{q}, \vec{o}_{i,<t})}}

\begin{align}
  \mathcal{J}_\text{GRPO}(\theta) &=  
  \mathbb{E}_{\vec{q}\sim\mathcal{Q},\{\vec{o}_i\}_{i=1}^G\sim\oldpi(\cdot|\vec{q})} \nonumber 
  \Bigg\{
  \frac{1}{G}\sum_{i=1}^G
  \frac{1}{|\vec{o}_i|} \sum_{t=1}^{|\vec{o}_i|}
  \min\Bigg[\piratio,
  \\& \text{clip}\left(\piratio,1-\epsilon,1+\epsilon\right) 
   \Bigg]
  - \beta \mathbb{D}_\text{KL}\left[\pi_\theta\|\pi_\text{ref}\right]
  \Bigg\},\label{eq:grpo}
\end{align}
where $\pi_\theta$ is the policy model, i.e., VLM.
For a query $\vec{q}$, we sample $G$ outputs $\{\vec{o}_1, \dots, \vec{o}_G\}$ from the old policy model $\oldpi$.
Clip ratio $\epsilon$ and $\beta$ are hyper-parameters and $\pi_\text{ref}$ is the reference VLM. 

\section{Hyperparameters}
\label{sec:hyperparameters}

In this section, we discuss the hyperparameters used in our SFT and RL experiments.

\subsection{Supervised Fine-Tuning}

For the SFT experiments, we use the AdamW optimizer~\citep{adamw19} with a learning rate of $1\times 10^{-5}$,  weight decay of 0.0, $\beta_1=0.9$ and $\beta_2=0.999$.
We adopt a multiplicative learning rate scheduler with a decay rate of 0.8 for each epoch. 
The training batch size is set as 4 per device and we used 32 GPUs in total.
The context length is truncated to 4096 tokens for each sequence.
Each dataset will be trained for 4 epochs.

\subsection{Reinforcement Learning}

Throughout our experiments, we set KL coefficient $\beta = 1.0 \times 10^{-2}$ and the clip ratio $\epsilon = 0.2$. 
For each example, we sample 8 responses with a maximum length of 4096 tokens and sampling temperature of 1.0. 

When updating the actor model, we use
AdamW~\citep{adamw19} and set the learning rate as $3.0\times 10^{-6}$ and weight decay as $1.0\times 10^{-2}$. 
A warm-up learning rate scheduler is used by using 10\% of the total training steps.
Gradient norm is applied with a threshold of 1.0.
A global batch size (calculated across all devices) is set as 128.
All the training are stopped on the convergence of the reward value on the training set.

\section{Prompt Templates}
\label{sec:prompts}
Below are the prompt templates in Jinja format~\cite{jinja} used in training and evaluation:

\begin{AIbox}{Prompt Template for Training}
\begin{verbatim}
You will solve a problem/request. You should provide your thoughts
within <think> </think> tags before providing the answer.\nWrite your
final answer within <answer> </answer> tags. Here is the question: 
{{ question }}{%
\end{verbatim}
\end{AIbox}

\begin{AIbox}{Prompt Template for Evaluation Multi-choice Questions}
\begin{verbatim}
You should provide your thoughts within <think> </think> tags, then
answer with just one of the options below within <answer> </answer> 
tags (For example, if the question is \n'Is the earth flat?\n A: Yes
\nB: No', you should answer with <think>...</think> <answer>B: No
</answer>). Here is the question: {{ question }}{%
Options:\n\n{{ options }}{%
\end{verbatim}
\end{AIbox}

\begin{AIbox}{Prompt Template for Evaluation Mathematical Questions}
\begin{verbatim}
A conversation between User and Assistant. The user asks a question,
and you as the assistant solves it. You should first think about 
the reasoning process in the mind and then provide the user with 
the answer. The reasoning process and answer are enclosed within 
<think> </think> and <answer> </answer> tags, respectively, i.e., 
<think> reasoning process here </think><answer> the final answer 
as the option letter or the number depending on the question 
</answer> (For example, if the question is \n'Is the earth flat?\n
A: Yes \nB: No', you should answer with <think> your reasoning 
</think> <answer>B: No</answer>. If the question is 'What is 1+1?',
you should answer with <think> your reasoning </think> <answer>2
</answer>).\n\nHere is the question: {{ question }}{%
Options:\n\n{{ options }}{%
\end{verbatim}
\end{AIbox}

\begin{table}[!ht]
\caption{Tasks for different evaluation settings.}
\label{tab:tasks}
\centering
\tablestyle{4pt}{1.2}
\begin{tabular}{@{}cllr@{}}
\shline
Evaluation setting & Task name & Shorthand Used & Data size \\ \hline
\multirow{2}{*}{General-domain text-only} & GSM8K main~\citep{cobbe2021training} & GSM8K & 1,319 \\
 & MMLU-Pro~\citep{wang2024mmlu} & MMLU-Pro & 12,032 \\ \hline
\multirow{5}{*}{\begin{tabular}[c]{@{}c@{}}General-domain multimodal\\ (X-modality)\end{tabular}} & MMMU (val)~\citep{yue2024mmmu} & MMMU & 900 \\
 & MMMU-Pro~\citep{yue2024mmmupro} & MMMU-Pro & 1,730 \\
 & MMStar~\citep{chenwe} & MMStar & 1,500 \\
 & MathVista (testmini)~\citep{lumathvista} & MathVista & 1,000 \\
 & MathVision~\citep{wang2024measuring} & MathVision & 3,040 \\ \hline
\multirow{3}{*}{\begin{tabular}[c]{@{}c@{}}Specialized-domain text-only\\ (X-domain)\end{tabular}} & MedQA~\citep{jin2021disease} & MedQA & 1273 \\
 & MedQA (4 options)~\citep{jin2021disease}& MedQA (4-ops) & 1,273 \\
 & MMLU-Pro-Health~\citep{wang2024mmlu} & MMLU-Pro-H & 818 \\ \hline
\multirow{5}{*}{\begin{tabular}[c]{@{}c@{}}Specialized-domain multimodal\\ (X-modality \& X-domain)\end{tabular}} & MMMU-Health~\citep{yue2024mmmu}& MMMU-H & 150 \\
 & MMMU-Pro-Health~\citep{yue2024mmmupro} & MMMU-Pro-H & 286 \\
 & MedXpertQA-MM~\citep{chenwe} & MedXpertQA-MM & 2,000 \\
 & OmniMedVQA~\citep{hu2024omnimedvqa} & OmniMedVQA & 1,000 \\ 
 & NEJM Image Challenge & NEJM & 947\\
 & \citep{nejm_image_challenge} &&\\
 \shline
\end{tabular}
\end{table}

\section{Evaluation Tasks}
\label{sec:tasks}
\Cref{tab:tasks} summarizes tasks used for different evaluation settings in our experiments: general-domain text-only, general-domain multimodal, specialized-domain text-only, and specialized-domain multimodal. We focus specifically on the medical domain as our representative specialized domain due to its significant practical importance and recent intensive research activities~\citep{llava-med,llava-rad,biomedclip,luo2022biogpt,medprompt,nori2024medprompt,codella2024medimageinsight,10.1145/3580305.3599568,xie2023doclens,ness2024medfuzzexploringrobustnesslarge}.

\section{Previous SOTA Results}

\begin{table}[!ht]
\caption{Previous SOTA results. * indicates our replicated results. \label{table:previous SOTA}}
\label{tab:previous sota}
\centering
\tablestyle{5pt}{1.3}
\begin{tabular}{@{}cll}
\shline
Task & Previous SOTA Model & Result \\
\midrule
GSM8K & MMOS-DeepSeekMath-7B \citep{chen2024empirical} & 87.2 \\
MMLU-Pro & Gemma-2-9B-it \citep{team2024gemma} & 52.1\\
MMMU-Pro & MAmmoTH-VL2-7B \citep{jia2025visualwebinstruct} & 40.7 \\
MMMU (Val) & Mulberry-7B \citep{yao2024mulberry} & 55.0 \\
MedQA (4-ops) & HuatuoGPT-o1-7b \citep{chen2024huatuogpt} & 71.6* \\
MMLU-Pro-H & HuatuoGPT-o1-7B \citep{chen2024huatuogpt} & 54.3* \\
MMMU-H & GMAI-VL-RL \citep{su2025gmai} & 57.3 \\
MedXpertQA-MM & GMAI-VL-Rl \citep{su2025gmai} & 23.8 \\
\shline
\end{tabular}
\end{table}

\section{Comparison of RL and SFT on MedQA data}
We conduct a direct comparison between Rl, SFT and their combination with the same MedQA data. As shown in \Cref{tab:rl vs sft}, pure RL performs worse than SFT and the hybrid SFT + RL performs the best.
\vspace{-1em}
\begin{table}[htbp]
\centering
\tablestyle{5pt}{1}
\caption{Comparison of performance from RL, SFT, and SFT + RL in both in-distribution (ind. text) and generalization settings, including out-of-distribution (ood. text) and cross-modality (X-modality). All models are trained on MedQA data.}
\label{tab:rl vs sft}
\begin{tabular}{lcccc}
\toprule
\multirow{2}{*}{Method} & MedQA & MMLU-Pro-Health  & MMMU-Pro-Health & Average Response Length \\
& (ind. text) & (ood. text) & (X-modality) & (number of words)\\
\midrule
baseline & 50.5 & 47.8 & 34.3 & 111\\ %
SFT & 61.9 & 54.2 & 39.2 &  950\\ %
RL & 57.8 & 53.6 & 35.8 & 103\\ %
SFT + RL & {\bf 64.9} & {\bf 55.8} & {\bf 39.7} & 1039\\ %
\bottomrule
\end{tabular}
\end{table}

\section{Training Dynamics}
\label{sec:dynamics}

\begin{figure}[!ht]
  \centering
  \begin{subfigure}[b]{0.33\textwidth}
    \centering
    \includegraphics[width=\linewidth]{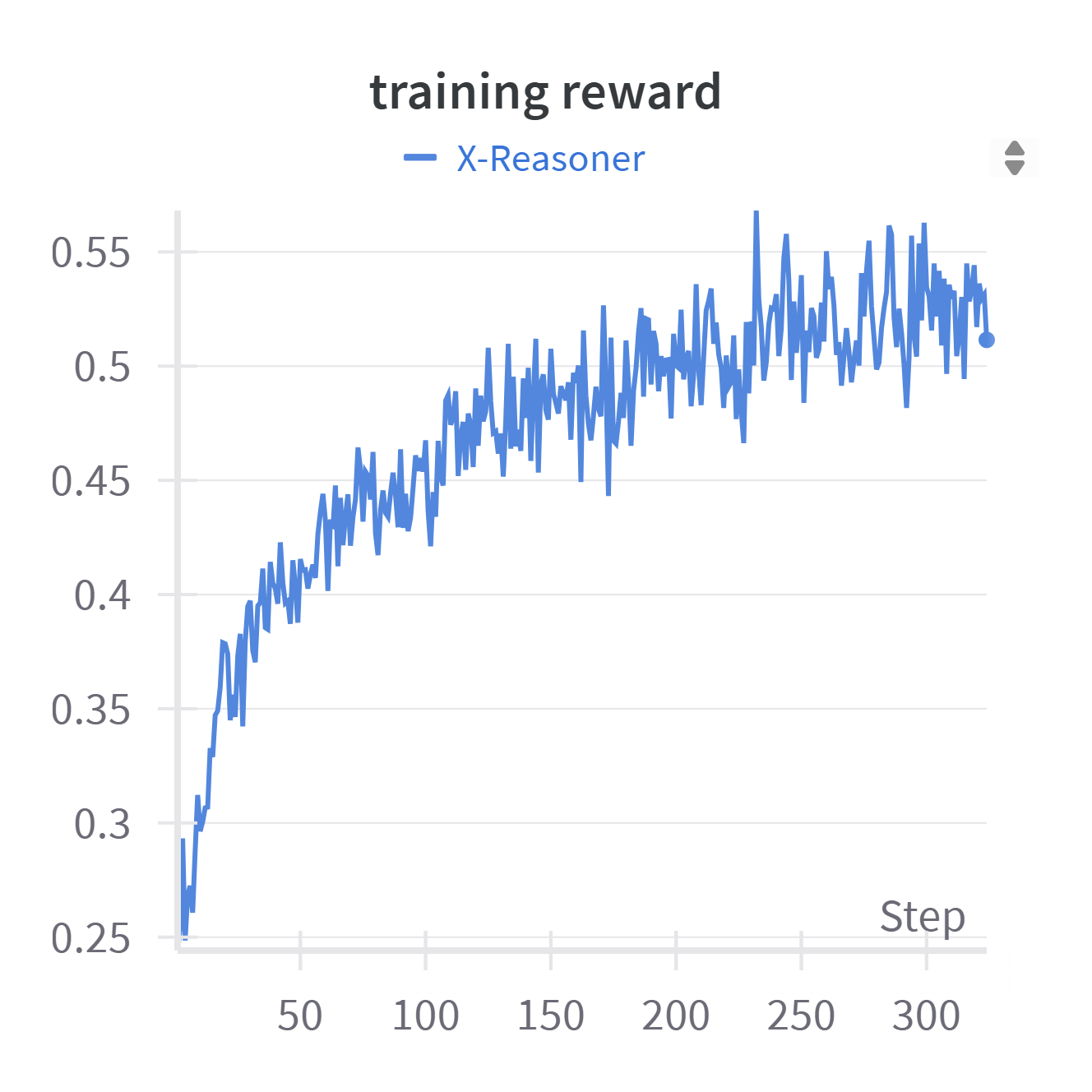}
  \end{subfigure}%
  \hfill
  \begin{subfigure}[b]{0.33\textwidth}
    \centering
    \includegraphics[width=\linewidth]{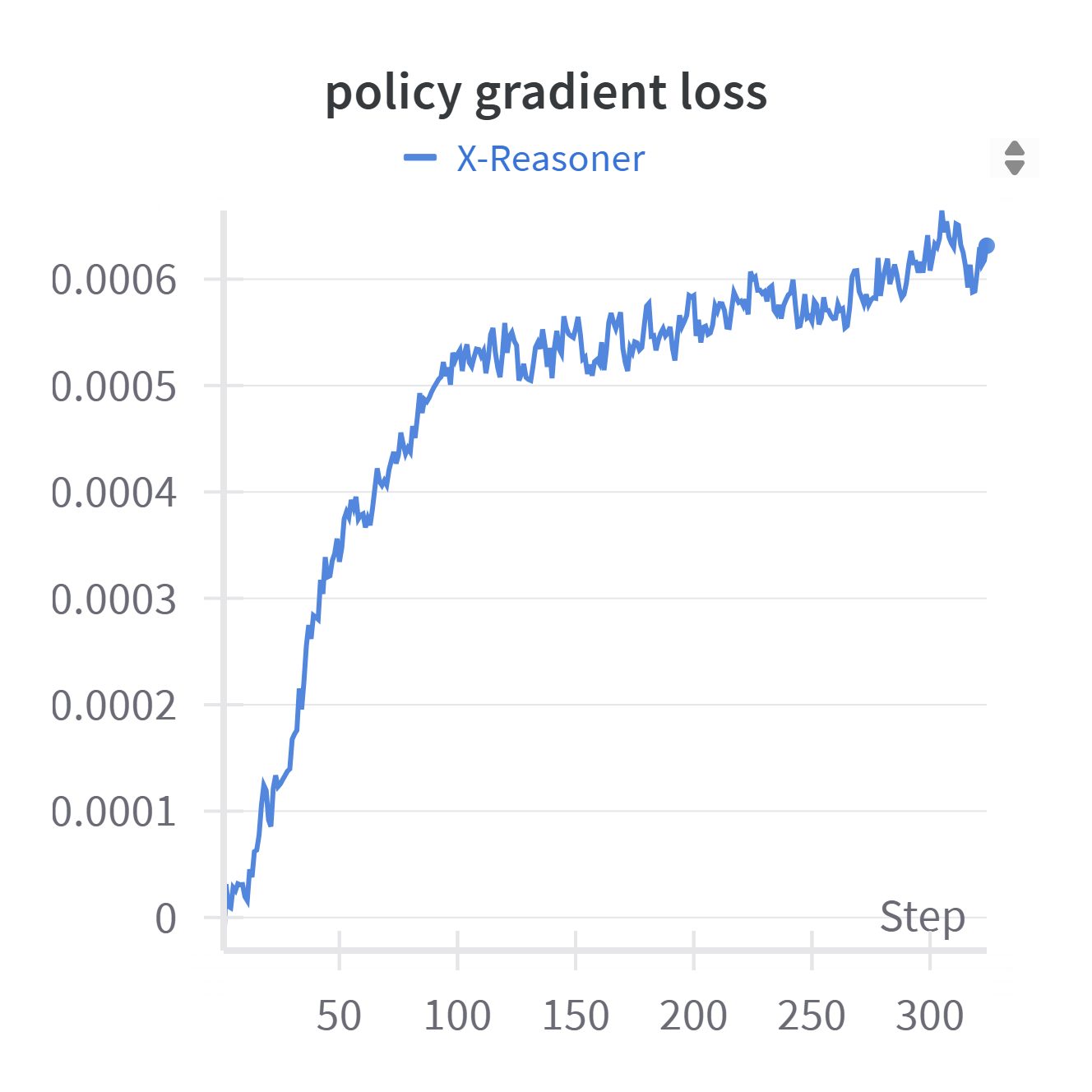}
  \end{subfigure}%
  \hfill
  \begin{subfigure}[b]{0.33\textwidth}
    \centering
    \includegraphics[width=\linewidth]{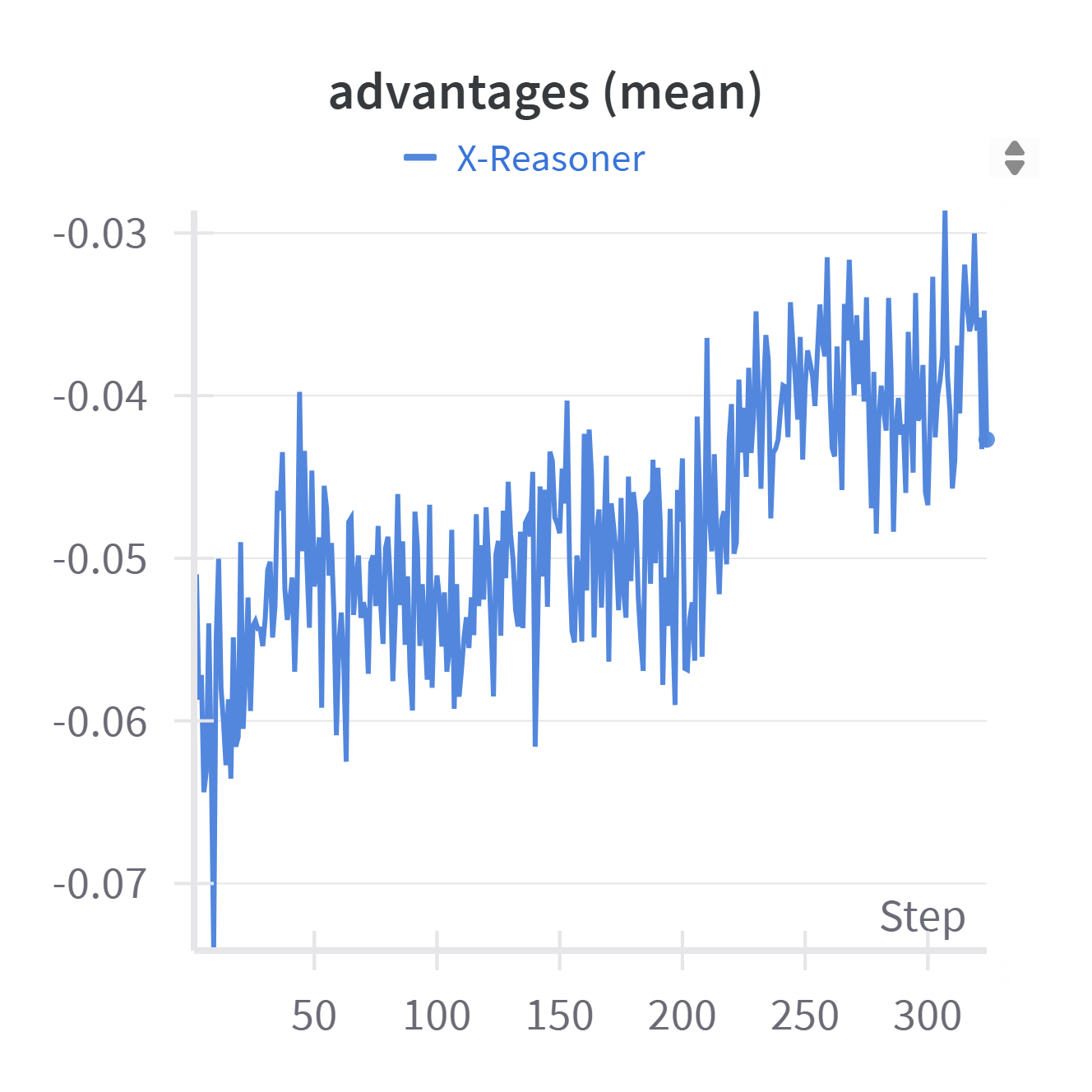}
  \end{subfigure}

  \begin{subfigure}[b]{0.33\textwidth}
    \centering
    \includegraphics[width=\linewidth]{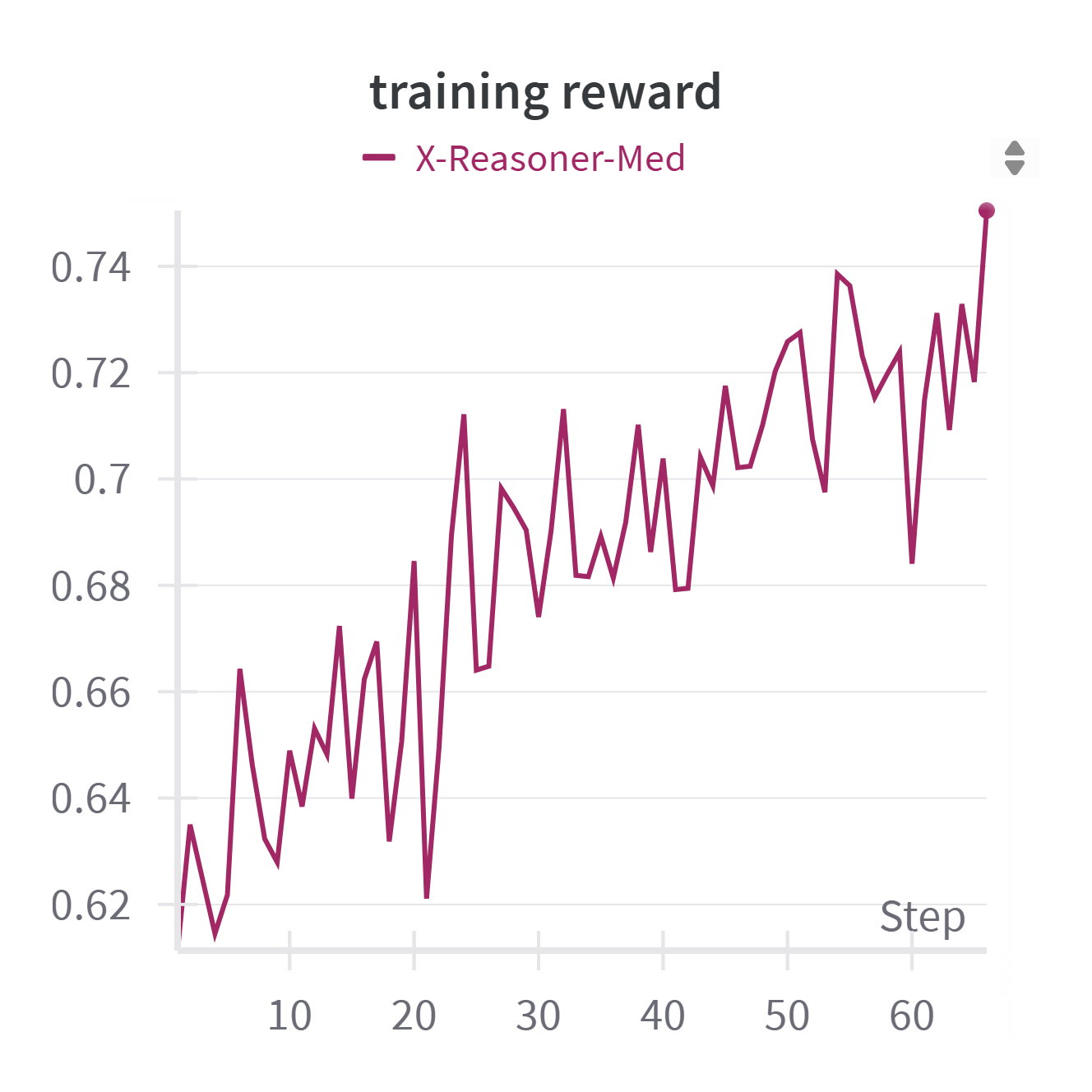}
  \end{subfigure}%
  \hfill
  \begin{subfigure}[b]{0.33\textwidth}
    \centering
    \includegraphics[width=\linewidth]{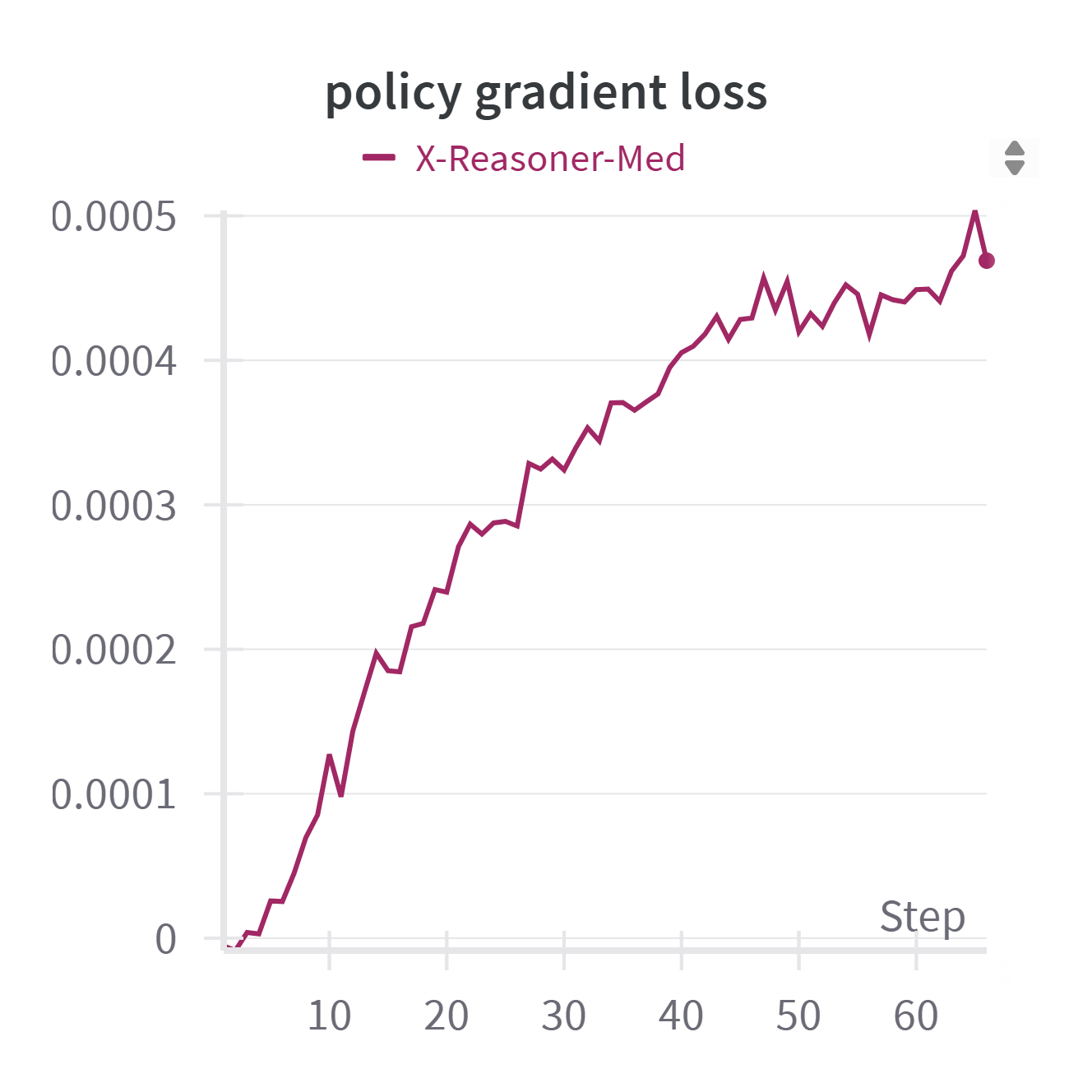}
  \end{subfigure}%
  \hfill
  \begin{subfigure}[b]{0.33\textwidth}
    \centering
    \includegraphics[width=\linewidth]{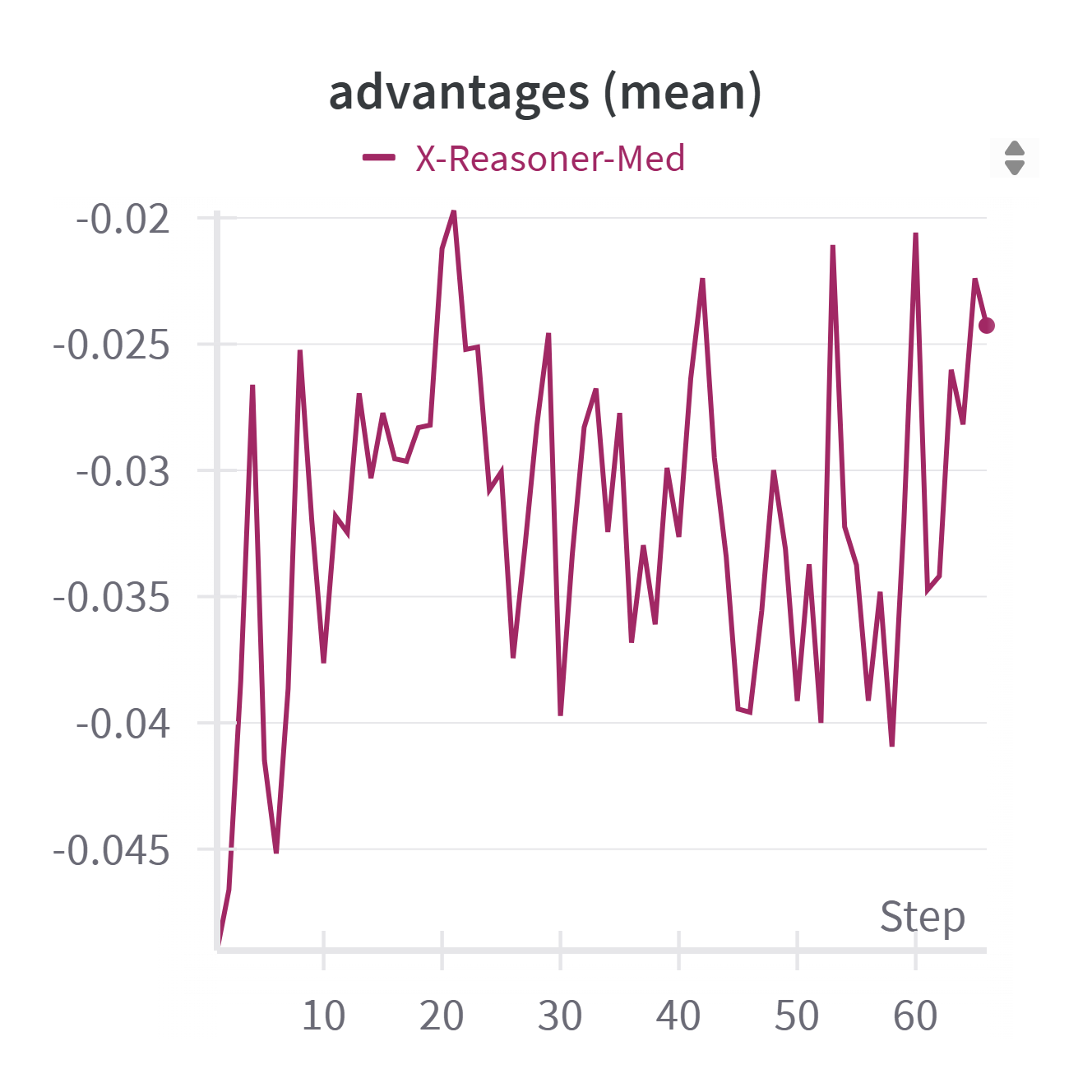}
  \end{subfigure}

  \begin{subfigure}[b]{0.33\textwidth}
    \centering
    \includegraphics[width=\linewidth]{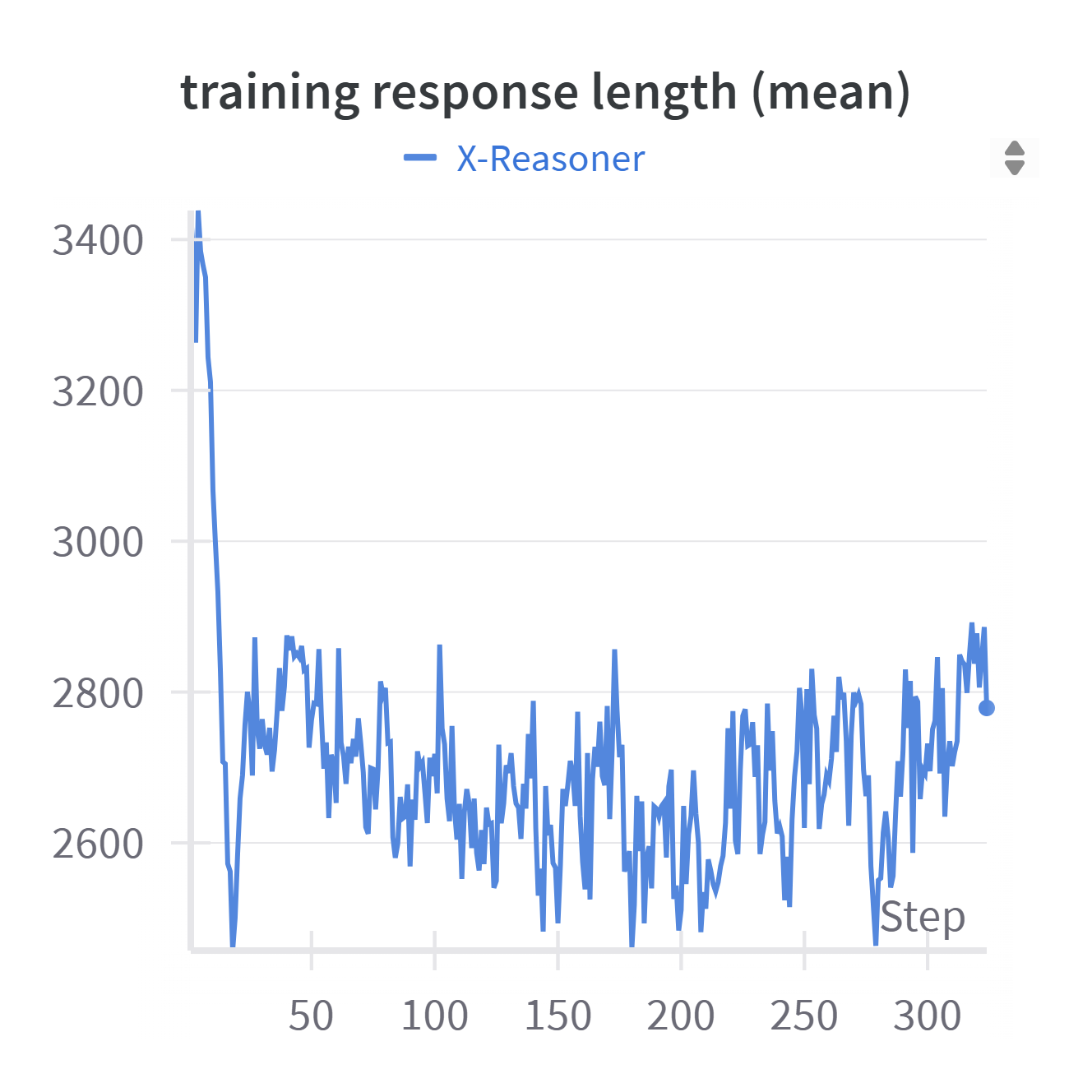}
  \end{subfigure}%
  \hfill
  \begin{subfigure}[b]{0.33\textwidth}
    \centering
    \includegraphics[width=\linewidth]{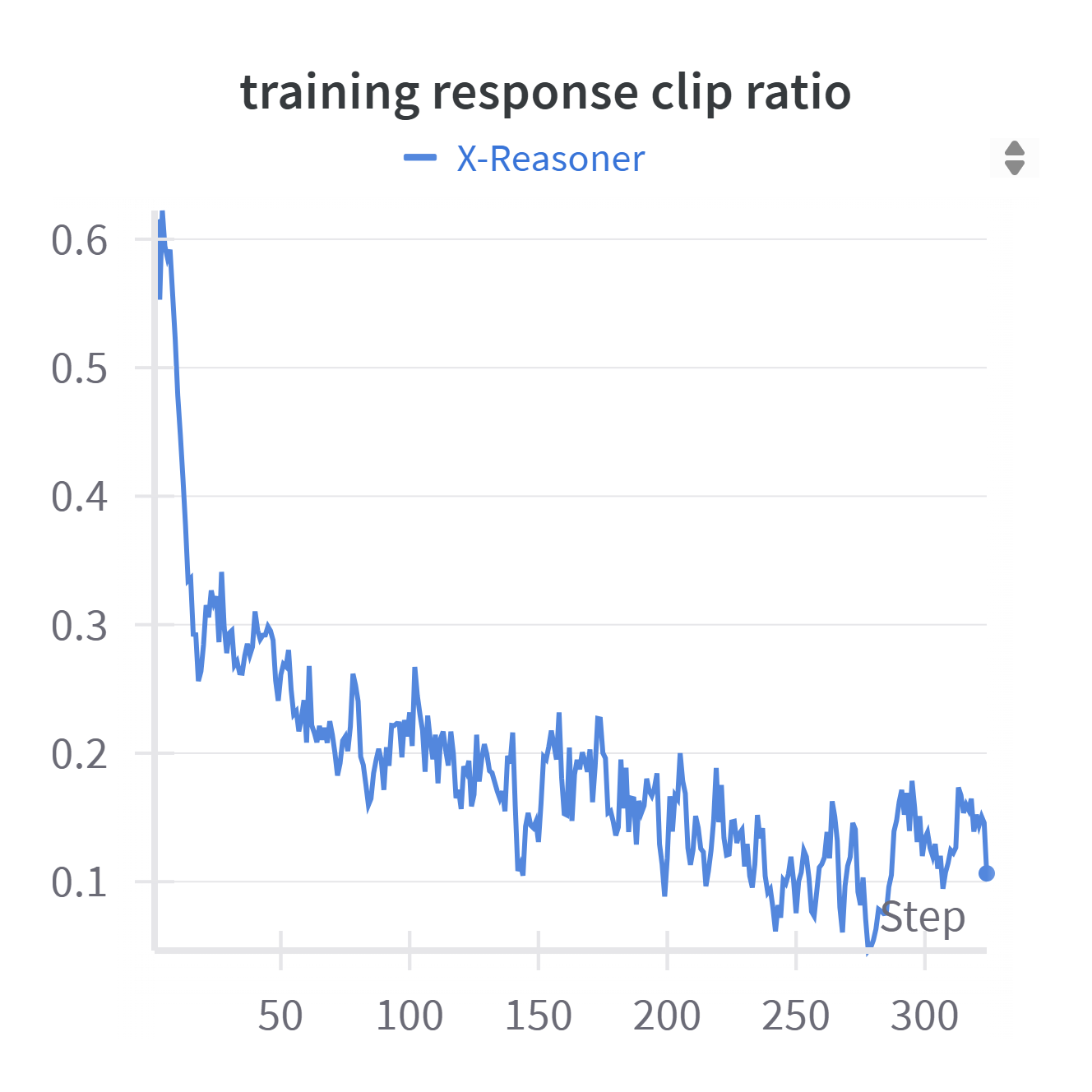}
  \end{subfigure}%
  \hfill
  \begin{subfigure}[b]{0.33\textwidth}
    \centering
    \includegraphics[width=\linewidth]{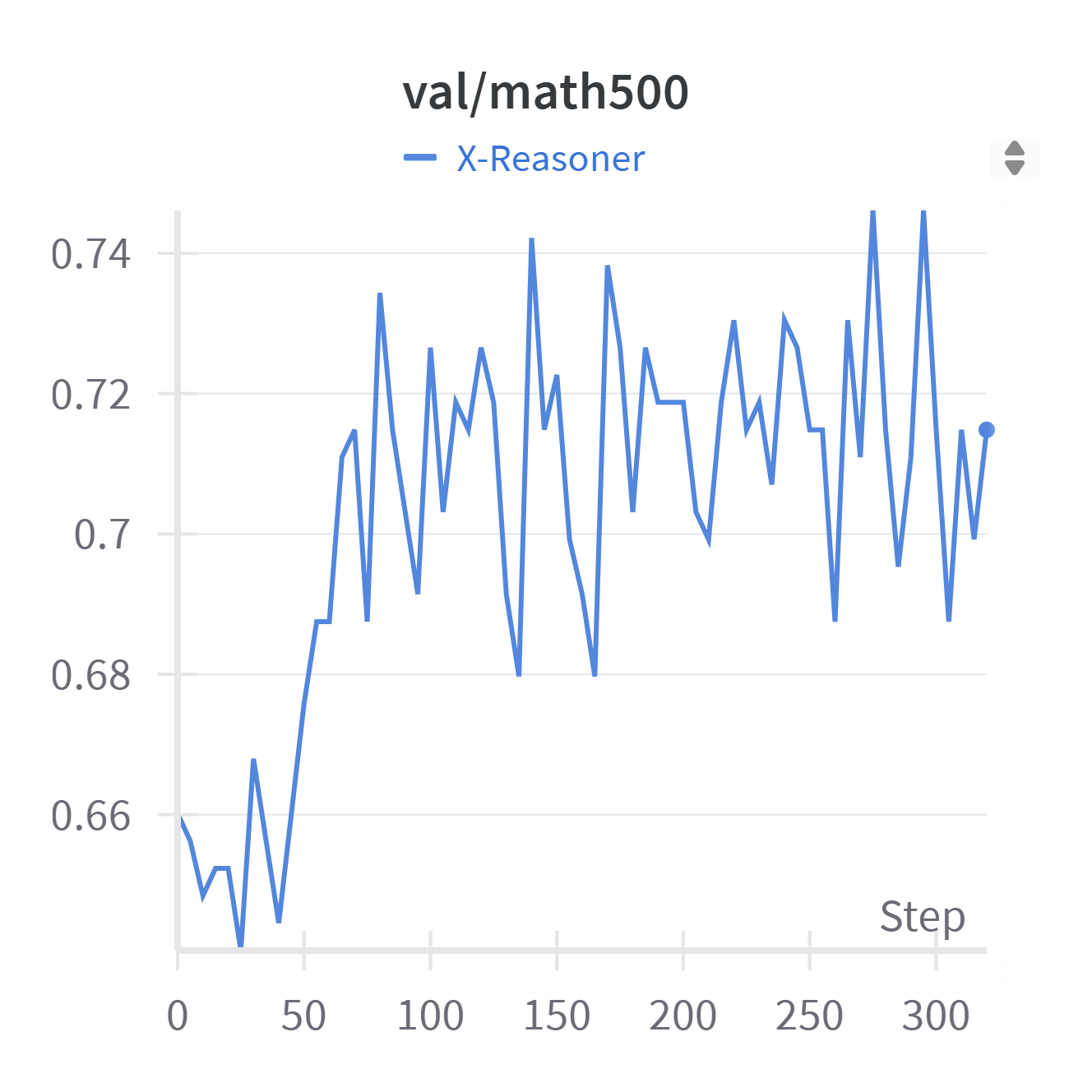}
  \end{subfigure}

  \begin{subfigure}[b]{0.33\textwidth}
    \centering
    \includegraphics[width=\linewidth]{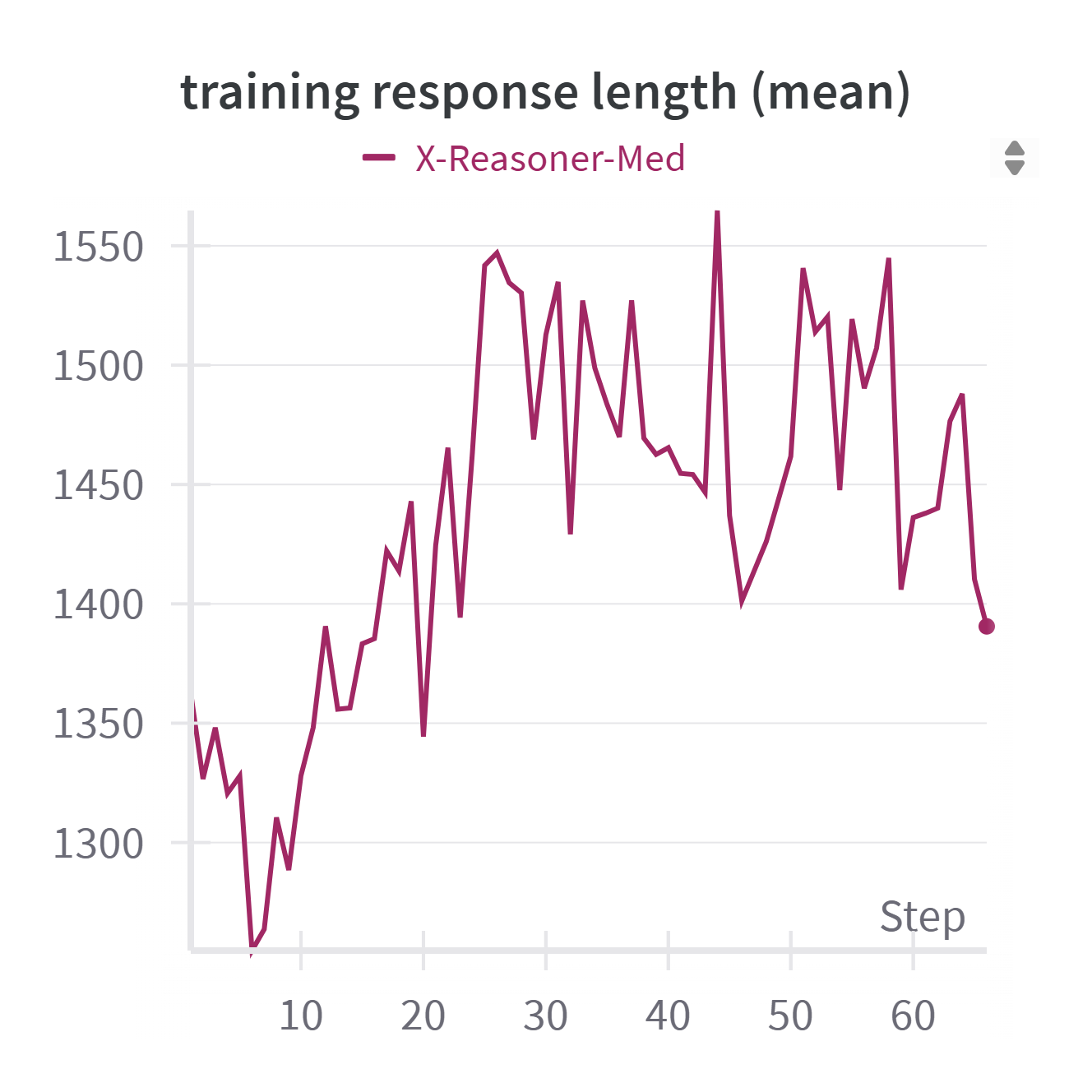}
  \end{subfigure}%
  \hfill
  \begin{subfigure}[b]{0.33\textwidth}
    \centering
    \includegraphics[width=\linewidth]{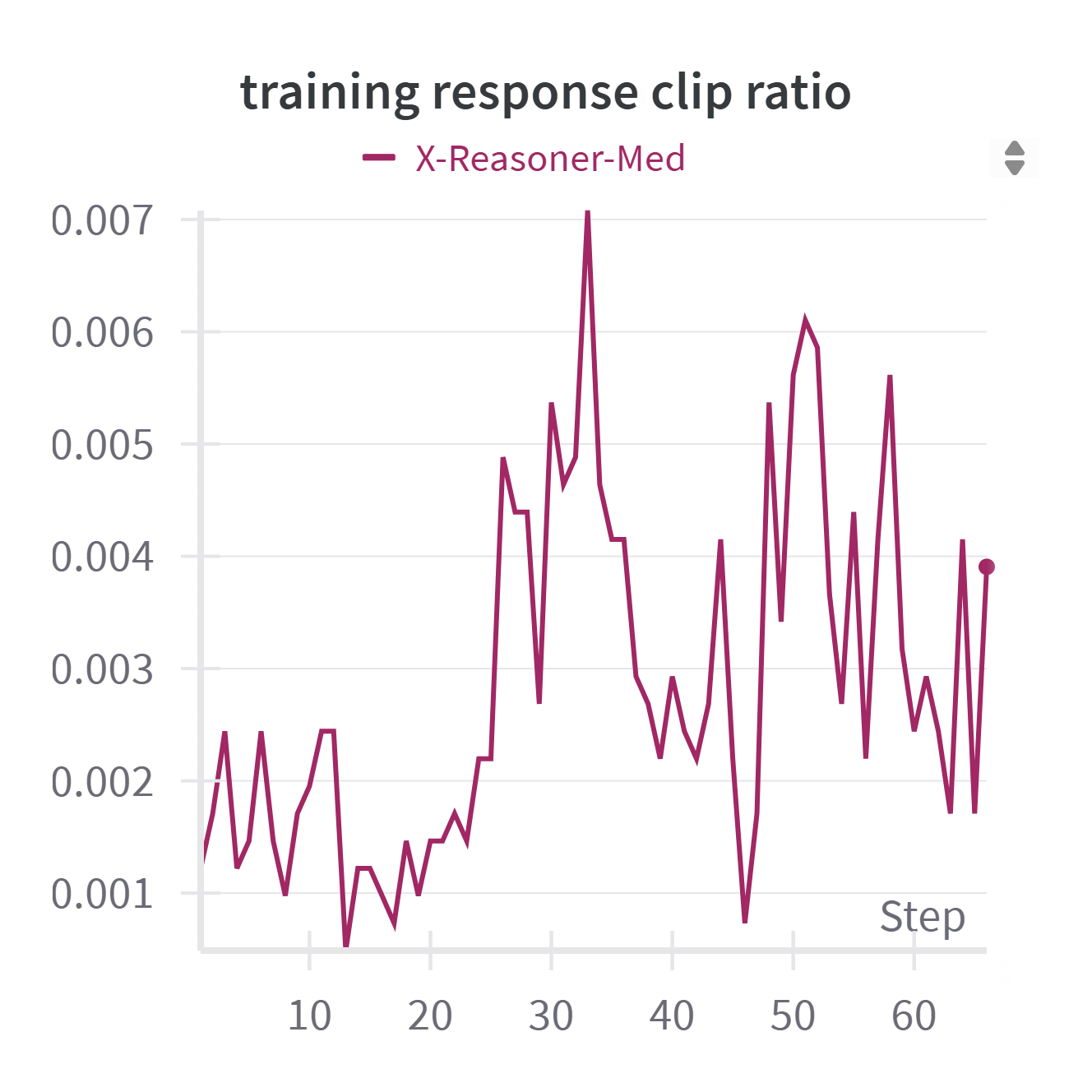}
  \end{subfigure}%
  \hfill
  \begin{subfigure}[b]{0.33\textwidth}
    \centering
    \includegraphics[width=\linewidth]{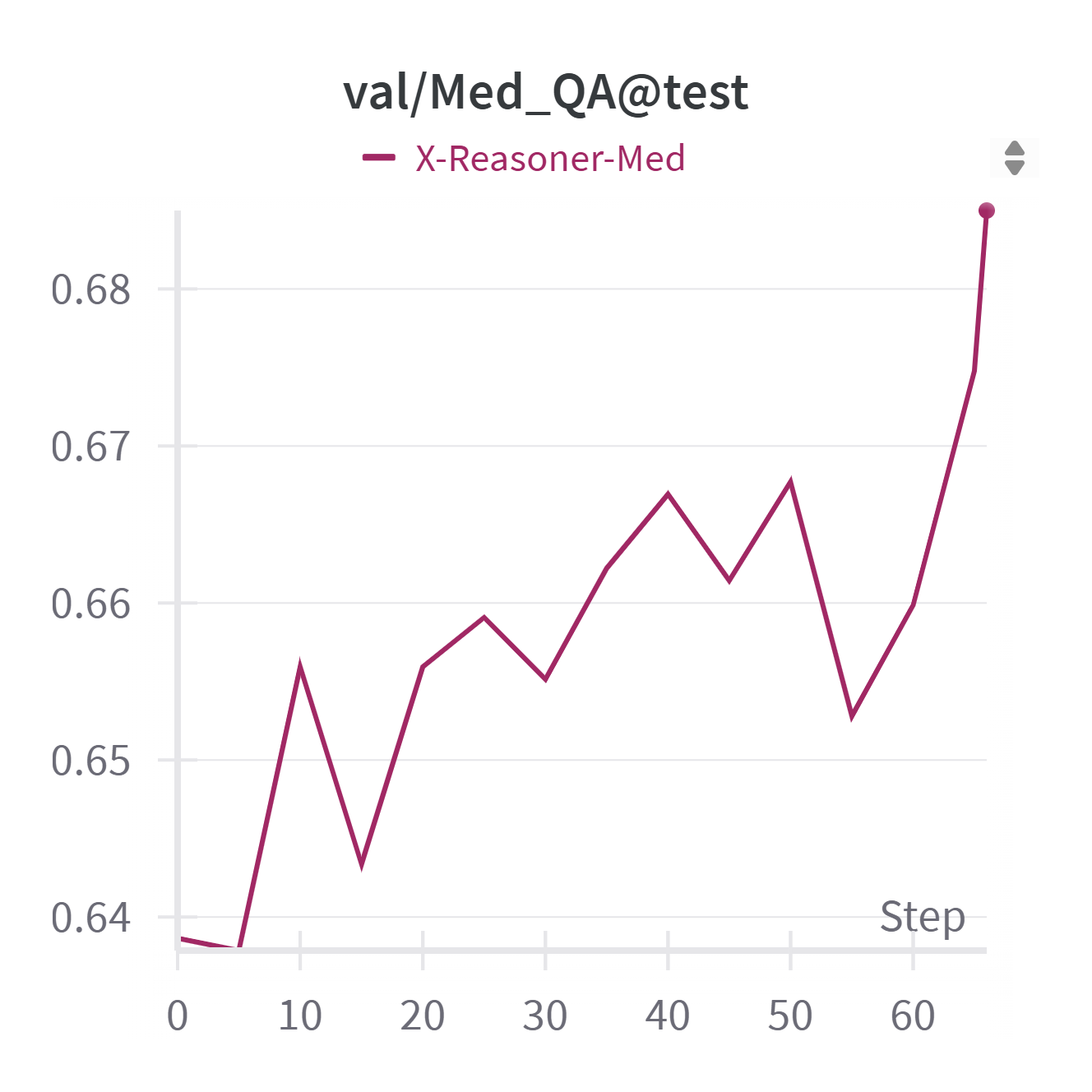}
  \end{subfigure}
  \caption{%
  RL training dynamics of \ours and \oursmed
  }
  \label{fig:dynamics}
\end{figure}

\Cref{fig:dynamics} shows the RL training dynamics when \ours is being trained on Orz~\citep{hu2025openreasonerzeroopensourceapproach} and when \oursmed is being trained on MedQA~\citep{jin2021disease}, including training reward, policy gradient loss, mean advantage, mean training response length, training response clip ratio, validation accuracy.

\newpage
\section{Qualitative Analysis}
\label{sec:examples}
\Cref{tab:text-example} shows some example reasoning output from \ours as compared with the baseline. We can clearly see that \ours is able to engage in more complex reasoning traces including detailed planning, verification ("wait). This complex reasoning patterns is successfully transferred to cross-modality setups where 
\ours seamlessly integrates the vision information into the reasoning chain, and provides detailed step-by-step reasoning with verification for review/correction. In the example, \ours provides a detailed plan to go over all the bar values whereas the baseline misses half of the bar values. \ours is also able to correct its mistake in the reasoning chain. In the cross-modal cross-modality setting, \ours connects the dots from both visual information and domain knowledge to the reasoning chain, and successfully identifies the patterns (white matter) from the histology slide and refers to its medical domain knowledge to connect it to progressive multifocal leukoencephalopathy.
\begin{table}[ht!]
\centering
\tablestyle{5pt}{1.3}
\caption{Reasoning output comparison between the baseline \base\xspace model and \ours on an text-only question.}
\label{tab:text-example}
\begin{tabular}{p{2cm}p{11cm}}
\toprule
\textbf{Question} & The great dragon, Perg, sat ...within a distance of 1000 feet.  Polly could throw the gold javelin...for a distance of 400 feet, well within the reach of the dragon's flames.  But when Polly held the sapphire gemstone, she could throw the javelin three times farther than when not holding the gemstone. If holding the gemstone, how far outside of the reach of the dragon's flames could Polly stand and still hit the dragon with the gold javelin? (from gsm8k main) \\
\midrule
\midrule
& \textbf{Response} \\
\midrule
\textbf{Baseline} & \verb|<think>| To solve this problem, we need to determine the maximum distance Polly can stand from the dragon and still hit it with the javelin when holding the sapphire gemstone. Here's the step-by-step reasoning:
1. When Polly is not holding the gemstone, she can throw the javelin for a distance of 400 feet.
2. When Polly holds the gemstone, she can throw the javelin three times farther than when not holding the gemstone. Therefore, the distance she can throw the javelin when holding the gemstone is \verb|\( 400 \times 3 = 1200 \)| feet...\verb|</think>|\newline\verb|<answer>|1200\verb|</answer>|  \xredmark \\
\hline
\textbf{\ours} & \verb|<think>|Okay, let's see. {\bf First, let's break down the information.} Without the gemstone, the javelin's distance is 400 feet. With the gemstone, it's three times that. So, 400 feet multiplied by 3 gives the new throwing distance. Let me calculate that: 400 * 3 = 1200 feet... {\bf Wait}, but is that correct? Let me check again. Wait, the problem says "how far outside of the reach of the dragon's flames could Polly stand and still hit the dragon with the gold javelin?" So, if the dragon's flames reach 1000 feet, then the maximum distance Polly can stand from the dragon's position and still hit it is 1200 feet. But the question is asking how far outside the dragon's flames she can stand. So, 1200 feet minus 1000 feet is 200 feet. Therefore, the answer should be 200 feet.
\verb|</think>|
\verb|<answer>200</answer>| \cgreenmark\\
\bottomrule
\end{tabular}
\end{table}

\begin{table}[ht!]
\centering
\tablestyle{5pt}{1.3}
\caption{Reasoning output comparison between the baseline \base\xspace model and \ours on a multimodal question.}
\label{tab:x-modality-example}
\begin{tabular}{p{2cm}p{11cm}}
\toprule
& \includegraphics[width=0.6\textwidth]{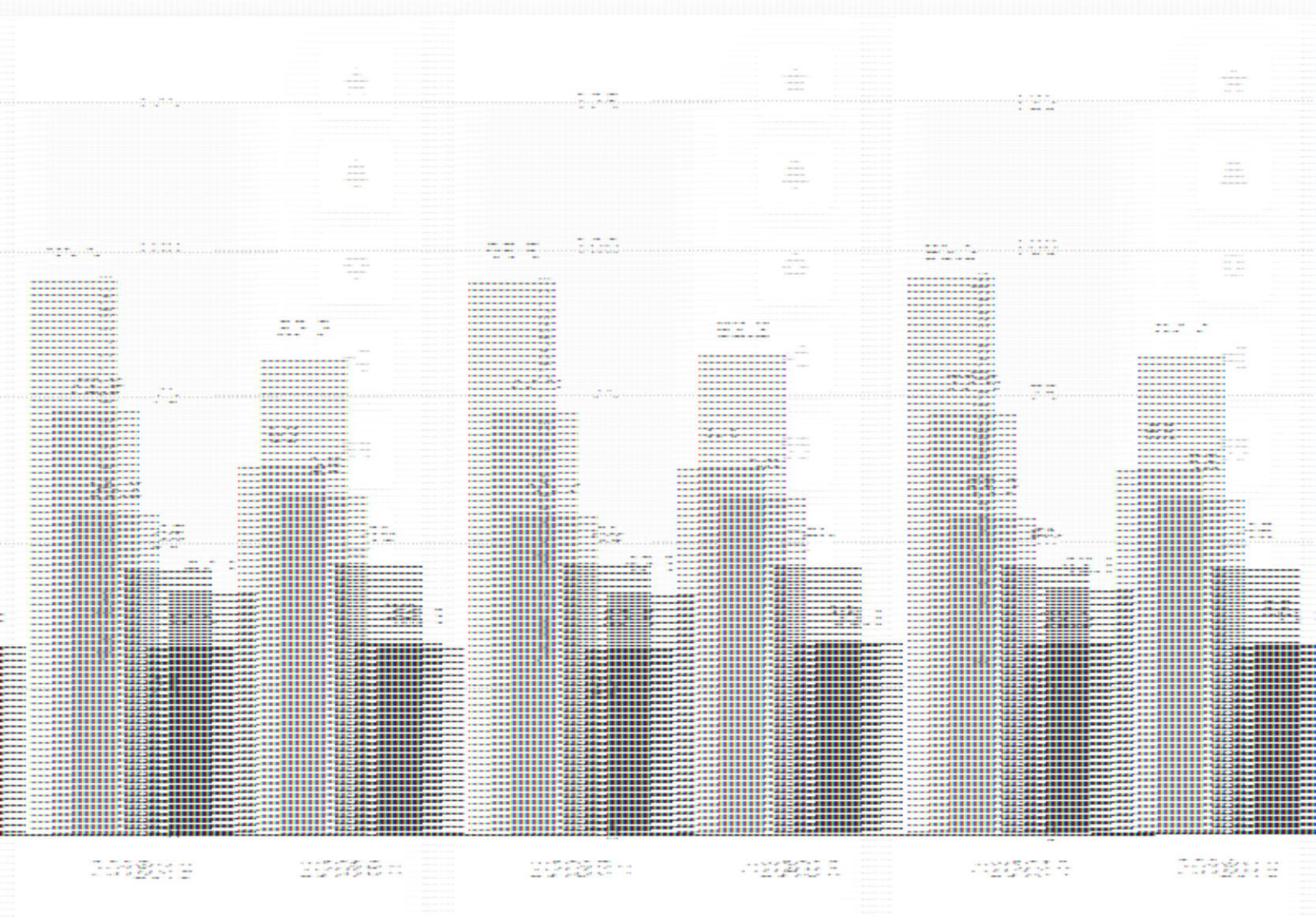} \\
\textbf{Question} & How many bars have value below 40? (from MathVista) \\
\midrule
\midrule
& \textbf{Response} \\
\midrule
\textbf{Baseline} &
\verb|<think>|To determine how many bars have a value below 40, I'll examine each bar's value:\newline

1. 2014: 63 (above 40)\newline
2. 2015: 95.5 (above 40)\newline
3. 2016: 82.2 (above 40)\newline
4. 2017: 72.9 (above 40)\newline
5. 2018: 58 (above 40)\newline
6. 2019: 55.2 (above 40)
\newline
None of the bars have a value below 40.\verb|</think><answer>0</answer>|  \xredmark \\
\hline
\textbf{\ours} & 
\verb|<think>| Okay, let's see. The question is asking how many bars have a value below 40. The image is a bar chart with two categories: Banknotes and Coins, and years from 2014 to 2019. Each year has two bars, one for each category. The values are given in thousands, so we need to check each bar's value and count those below 40 thousand. {\bf First, let's go through each year}:\newline
2014:\newline
- Banknotes: 63 thousand. That's above 40.\newline
- Coins: 46 thousand. Also above 40.\newline
...\newline
Now, let's count the bars below 40:\newline
- 2016: Coins (33) \newline
- 2017: Coins (32.5) \newline
- 2018: Coins (33.1)\newline
- 2019: Coins (42.1) - {\bf Wait}, 42.1 is above 40. Wait, hold on. Wait, 2019's Coins is 42.1 thousand. 42.1 is above 40. So that's not below 40. So only the 2016 and 2017 Coins are below 40...
Let's recast:\newline

2014: Banknotes 63 (no), Coins 46 (no) → 0\newline
2015: Banknotes 95.5 (no), Coins 46 (no) → 0\newline
2016: Banknotes 82.2 (no), Coins 33 (yes) → 1\newline
2017: Banknotes 72.9 (no), Coins 32.5 (yes) → 1\newline
2018: Banknotes 58 (no), Coins 33.1 (yes) → 1\newline
2019: Banknotes 55.2 (no), Coins 42.1 (no) → 0\newline
...
So total bars below 40: 3. Therefore, the answer is 3.

\verb|</think>|

\verb|<answer>3</answer>| \cgreenmark
\\
\bottomrule
\end{tabular}
\end{table}

\begin{table}[ht!]
\centering
\tablestyle{5pt}{1.3}
\caption{Reasoning output comparison between the baseline \base\xspace model and \ours on an medical multimodal question.}
\label{tab:x-modality-domain-example}
\begin{tabular}{p{2cm}p{11cm}}
\toprule
 & \includegraphics[width=0.7\textwidth]{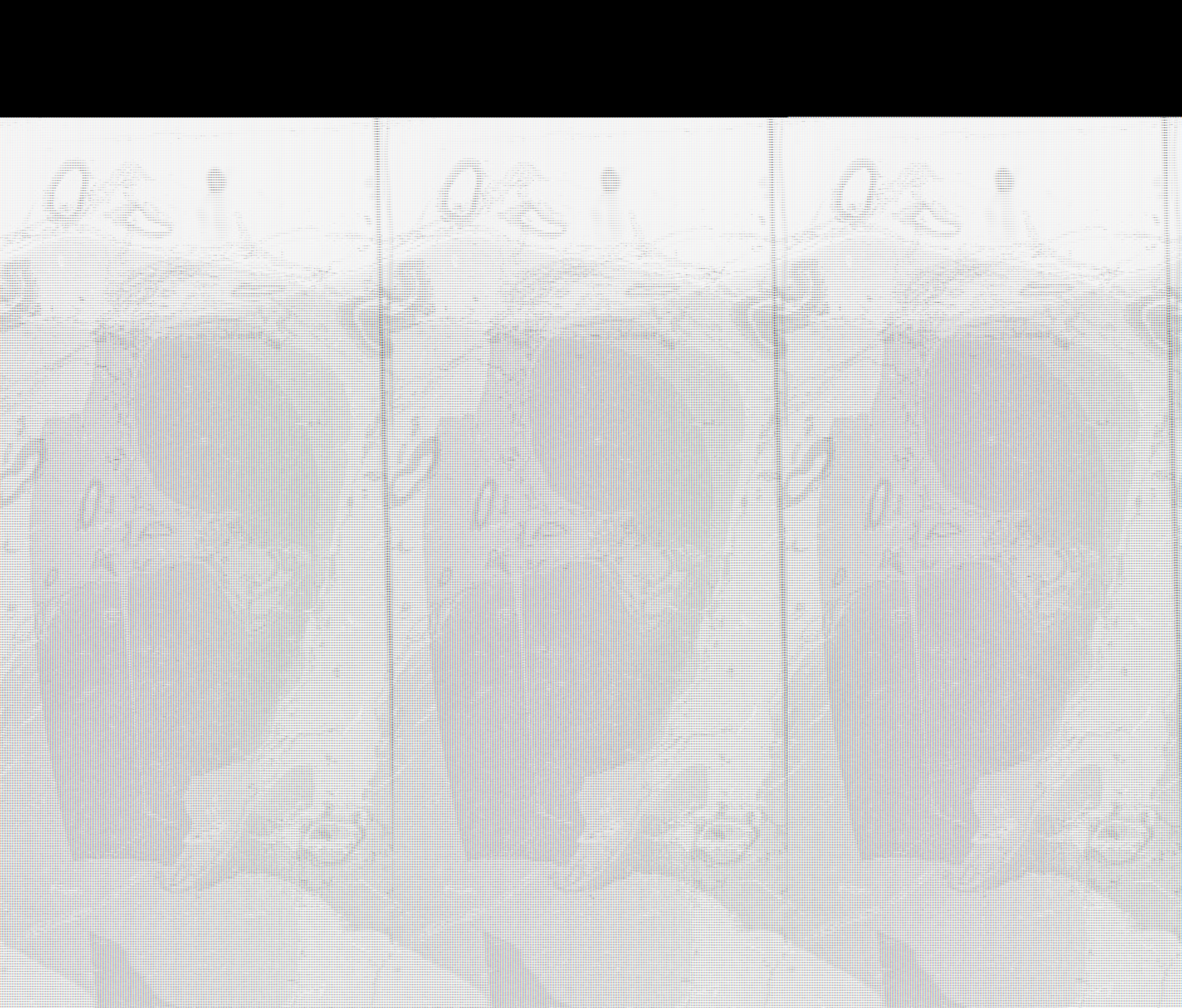} \\
\textbf{Question} & What is in the image? A. Progressive Multifocal Leukoencephalopathy (PML)
B. Rabies
C. West Nile Virus
D. Herpes Simplex encephalitis
E. Encephalitis Lethargica
F. HIV encephalitis
G. Lyme Disease
H. Syphilis
I. Meningitis
J. Creutzfeldt-Jakob disease (from MMMU\_PRO HEALTH)
\\
\midrule
\midrule
& \textbf{Response} \\
\midrule
\textbf{Baseline} &
\verb|<think>|The image appears to be a histological section, likely from a brain biopsy, given the presence of cellular structures and the context of the question. The pattern of the lesions and the cellular infiltration are characteristic of encephalitis. Herpes Simplex encephalitis (HSE) is a well-known cause of encephalitis, often presenting with a characteristic pattern of perivascular inflammation and necrosis, which can be seen in such images. The lesions in the image seem to have a focal nature, which is consistent with HSE. However, without further clinical information and a more detailed examination, a definitive diagnosis cannot be made.

Given the options, the most likely answer is:

D. Herpes Simplex encephalitis\verb|</think><answer>|D. Herpes Simplex encephalitis\verb|</answer>|  \xredmark
 \\
\hline
\textbf{\ours} & 
\verb|<think>|
Okay, let's try to figure out this question. The image shows a histological section with some lesions. The question is about which neurological disease is depicted. {\bf Let me go through the options one by one.}
First, I remember that PML (Progressive Multifocal Leukoencephalopathy) is caused by the JC virus and typically affects the white matter of the brain, leading to lesions that can be seen on MRI as areas of low signal intensity on T2-weighted images. The histology might show demyelination and gliosis. The image here shows some irregular areas with a pinkish hue, which could be consistent with demyelination. {\bf But I need to check} if the lesions are in the white matter. The image doesn't show the entire brain, but the lesions appear to be in the white matter, which is typical for PML.
Rabies usually presents with encephalitis, but the histology isn't typically as specific as PML... Encephalitis Lethargica is an old disease, but its histology isn't as specific as PML.

Given the appearance of the lesions in the white matter, the pinkish areas, and the possibility of demyelination, PML seems the most likely. So the answer is A.
\verb|</think>|

\verb|<answer>A</answer>| \cgreenmark
\\
\bottomrule
\end{tabular}
\end{table}

\end{document}